\documentclass[journal]{IEEEtran}

\usepackage{graphicx}
\usepackage{array}
\usepackage{xcolor}
\usepackage[colorlinks,
linkcolor=red,
anchorcolor=green,
citecolor=blue]{hyperref}
\usepackage{amsmath}
\usepackage{algorithm}
\usepackage{algorithmic}
\usepackage{ulem}
\usepackage{stfloats}
\usepackage{cite}
\usepackage{subfigure}
\usepackage{multirow}

\definecolor{dark-purple}{HTML}{8d4bbb}
\definecolor{dark-red}{HTML}{D63C3C}

\newcommand{\TODO}[1]{\textcolor{black}{#1}} 
\newcommand{\rOneFC}[1]{\textcolor{black}{#1}} 
\newcommand{\rOneMSB}[1]{\textcolor{black}{#1}}
\newcommand{\rOneSHK}[1]{\textcolor{black}{#1}}
\newcommand{\rTwoMSB}[1]{\textcolor{black}{#1}}
\newcommand{\rThreeMSB}[1]{\textcolor{black}{#1}}
\newcommand{\rTwoFC}[1]{\textcolor{black}{#1}}

\begin{document}

\setlength{\abovecaptionskip}{-1.5em}

\title{APT-LLM: Exploiting \underline{A}rbitrary-\underline{P}recision \\ \underline{T}ensor Core Computing for \underline{LLM} Acceleration}

% \author{Shaobo Ma$^1$, Chao Fang$^1$, Haikuo Shao$^1$, Zhongfeng Wang$^{1,2}$\\
% $^1$School of Electronic Science and Engineering, Nanjing University, Nanjing, China \\
% $^2$School of Integrated Circuits, Sun Yat-sen University, Shenzhen, China\\
% {shaoboma, fantasysee, hkshao}@smail.nju.edu.cn, zfwang@nju.edu.cn}

\author{Shaobo~Ma, Chao~Fang, Haikuo Shao, Zhongfeng~Wang,~\IEEEmembership{Fellow,~IEEE} 
\thanks{This work was supported in part by the National Key R\&D Program of China under Grant 2022YFB4400600 and in part by Postgraduate Research \& Practice Innovation Program of Jiangsu Province under Grant KYCX24\_0149. \textit{(Corresponding author: Chao Fang; Zhongfeng Wang.)}}
% \thanks{S. Ma, C. Fang, and H. Shao are with the School of Electronic Science and Engineering, Nanjing University (e-mail: shaoboma@smail.nju.edu.cn; fantasysee@smail.nju.edu.cn; hkshao@smail.nju.edu.cn).}% <-this % stops a space
\thanks{S. Ma, and H. Shao are with the School of Electronic Science and Engineering, Nanjing University (e-mail: shaoboma@smail.nju.edu.cn; hkshao@smail.nju.edu.cn).}% <-this % stops a space
\thanks{\rTwoFC{C. Fang was with the School of Electronic Science and Engineering, Nanjing University, and is now with Shanghai Qi Zhi Institute (email: fantasysee@smail.nju.edu.cn).}}
\thanks{Z. Wang is with the School of Electronic Science and Engineering, Nanjing University, and the School of Integrated Circuits, Sun Yat-sen University (email: zfwang@nju.edu.cn).}
% \thanks{J. Lin is with the School of Electronic Science and Engineering, Nanjing University, and the Interdisciplinary Research Center for Future Intelligent
% Chips (Chip-X), Nanjing University (email: jlin@nju.edu.cn).}
% \thanks{Manuscript received April 19, 2005; revised August 26, 2015.}
}

\maketitle

\begin{abstract}
    % Large language models (LLMs) have been widely applied but face challenges in efficient inference. 
    \rOneFC{Large language models (LLMs) have revolutionized AI applications, yet their enormous computational demands severely limit deployment and real-time performance.}
    % While quantization methods reduce computational demands, ultra-low bit quantization with arbitrary precision is hindered by limited GPU Tensor Core support and inefficient memory management, leading to suboptimal acceleration. 
    % \rOneFC{Quantization methods can help reduce computational costs, but achieving the extreme efficiency of ultra-low-bit quantized LLMs with arbitrary precision remains challenging on GPUs due to limited GPU Tensor Core support, inefficient memory management, and rigid kernel optimizations, resulting in suboptimal acceleration.}
    \rTwoMSB{Quantization methods can help reduce computational costs, however, attaining the extreme efficiency associated with ultra-low-bit quantized LLMs at arbitrary precision presents challenges on GPUs. This is primarily due to the limited support for GPU Tensor Cores, inefficient memory management, and inflexible kernel optimizations.}
    To tackle these challenges, we propose a comprehensive acceleration scheme for arbitrary precision LLMs\rOneFC{, namely APT-LLM}.
    %Fundamentally, we present an innovative bipolar-INT data format designed to enhance parallel processing and enable symmetric quantization, thereby diminishing data redundancy. Subsequently, we develop a matrix multiplication (MatMul) method allowing for arbitrary precision by dismantling and reassembling matrices at the bit level. This technique offers flexible precision and optimizes the utilization of GPU Tensor Cores. 
    Firstly, we introduce a novel data format, bipolar-INT, which allows for efficient and lossless conversion with signed INT, while also being more conducive to parallel computation. We also develop a matrix multiplication (MatMul) method allowing for arbitrary precision by dismantling and reassembling matrices at the bit level. This method provides flexible precision and optimizes the utilization of GPU Tensor Cores. 
    In addition, we propose a memory management system focused on data recovery, which strategically employs fast shared memory to substantially increase kernel execution speed and reduce memory access latency.
    Finally, we develop a kernel mapping method that dynamically selects the optimal configurable hyperparameters of kernels for varying matrix sizes, enabling optimal performance across different LLM architectures and precision settings.
    % \rOneMSB{For arbitrary-precision MatMul tasks, our method significantly outperforms NVIDIA CUTLASS and state-of-the-art acceleration techniques on an NVIDIA RTX 3090. In LLM inference, our approach achieves substantial speedups in both the prefill and decode stages, surpassing CUTLASS INT4 by up to 1.98$\times$ and 2.16$\times$, respectively.}
    % In LLM inference, APT-LLM achieves up to a 3.99$\times$ speedup compared to FP16 baselines and a 2.16$\times$ speedup over NVIDIA CUTLASS INT4 acceleration.
    \rThreeMSB{In LLM inference, APT-LLM achieves up to a 3.99$\times$ speedup compared to FP16 baselines and a 2.16$\times$ speedup over NVIDIA CUTLASS INT4 acceleration on RTX 3090. On RTX 4090 and H800, APT-LLM   achieves up to 2.44$\times$ speedup over FP16 and 1.65$\times$ speedup over CUTLASS integer baselines.}
\end{abstract}

\section{Introduction}\label{sec:intro}
% 【摘要+引言：1.5页】

% \begin{itemize}
%     \item 介绍LLM是什么以及LLM推理的难点

\IEEEPARstart{R}{ecent} advancements in large language models (LLMs), such as GPT~\cite{gpt-4},
% ~\cite{gpt-1,gpt-2,gpt-3,gpt-4}, 
LLaMA~\cite{llama-3},
% ~\cite{llama-1,llama-2,llama-3}, 
and Deepseek~\cite{deepseekv3,deepseekr1}, have shown remarkable capabilities across 
% diverse natural language processing (NLP) benchmarks~\cite{nlp}, % [Chao] 有些局限这个说法，LLM 的应用范围远超 NLP
\rOneFC{a wide spectrum of language understanding and generation tasks~\cite{nlp},}
with performance improving alongside the increase in model parameters~\cite{scaling-law,chateda}. 
However, the growth in model size presents significant challenges for computation and storage. 
As a result, reducing the computational and storage demands of LLMs has become a primary research focus. 
Among various optimization techniques, model quantization~\cite{yang2022dtatrans,qlora,gptq,huang2024precision,tsld,onebit,smoothquant,omni,liu2024holes} is one of the most effective solutions, as it compresses storage bit-width and lowers computational requirements.

    % \item 引出LLM量化推理的意义, 介绍已有的低比特量化LLM

% With the development of quantization algorithms, LLM quantization has reached ultra-low bit-widths, such as GPTQ\cite{gptq}, OmniQuant\cite{omni}, and OneBit\cite{onebit}, significantly reducing the memory footprint and computational requirements for model inference. 
\rOneFC{With developments in quantization algorithms like GPTQ\cite{gptq}, QuanDCL\cite{QuanDCL}, OmniQuant\cite{omni}, Atom\cite{atom}, and OneBit\cite{onebit}, LLMs can now operate at ultra-low bit-widths, significantly reducing memory footprint and computational requirements for inference.}
% \sout{However, these advanced quantization methods are not well-matched with the capabilities of mainstream GPU devices, leading to several challenges in practical deployment, primarily in the following aspects:}
\rOneMSB{However, these advanced quantization methods do not align well with the capabilities of mainstream GPU devices, mainly in the following aspects:}
    % \item 提出目前低比特量化LLM推理遇到的问题

\begin{enumerate}
    \item \textbf{\rOneMSB{At the format level: }Limited data format support in GPU Tensor Cores (TCs).} TCs are key hardware components in modern GPUs that accelerate matrix multiplication (MatMul), playing a dominant role in LLM inference\cite{a100}. Starting from the Turing architecture\cite{Turing}, NVIDIA's GPU Tensor Cores support certain low-precision tensor operations, such as INT1, INT4, and even bit-level operations\cite{tc_benchmark}. However, TCs still lack support for some data formats widely used in ultra-low-bit quantized LLMs, such as INT2 and INT3\cite{omni}. When deploying these models on GPUs, the low-bit quantized data needs to be converted into higher-bit data formats supported by TCs for computation\cite{gptq, omni, onebit}. This data format conversion introduces additional computational overhead, preventing some ultra-low-bit quantized LLMs from achieving optimal inference acceleration on GPU platforms.

    \item \textbf{\rOneMSB{At the memory level: }Inefficient GPU memory management schemes.} When deploying LLMs on GPUs, performance bottlenecks arise not only from MatMuls but also from data access\cite{qserve}. GPUs utilize a multi-level memory hierarchy, with each level offering varying capacities and bandwidths\cite{tc_benchmark}. Effective memory management is crucial for minimizing data access latency. For instance, by strategically leveraging different memory levels, such as texture memory and shared memory, memory-bound applications can achieve speedups of up to 14.8$\times$ compared to unoptimized implementations\cite{gblastn}. Hence, optimization schemes that focus exclusively on MatMuls, while neglecting GPU memory management, often fail to deliver the expected performance improvements. In some cases, they may even lead to slower execution times than the original unoptimized implementation\cite{arif2023accelerating,wang2018superneurons}.

    \item \textbf{\rOneMSB{At the kernel level: }\rOneFC{Rigid kernel optimization for diverse matrices.}% Inapplicability of a single kernel.
    } LLMs rely on MatMuls for key operations, but these computations often involve matrices of varying sizes and characteristics\cite{llmint8}. \rOneMSB{In Fig. \ref{fig:matmul different}, using the LLaMA3-8B model as an example, it is evident that the MatMul sizes vary significantly across different layers.} The challenge arises because many GPU kernels are optimized for a specific matrix size, usually the largest matrices involved\cite{quantllm}. While this can improve performance for those specific cases, it is not efficient when applied to smaller or larger matrices. Therefore, using a single GPU kernel for all matrix operations leads to suboptimal performance, as it fails to fully utilize GPU resources for diverse matrix sizes\cite{rivera2021tsm2x,osama2023stream}. This inefficiency not only increases memory usage but also introduces computational bottlenecks, ultimately slowing down inference times.
   
    \begin{figure}[t]
        \centering
        \includegraphics[width=0.9\linewidth]{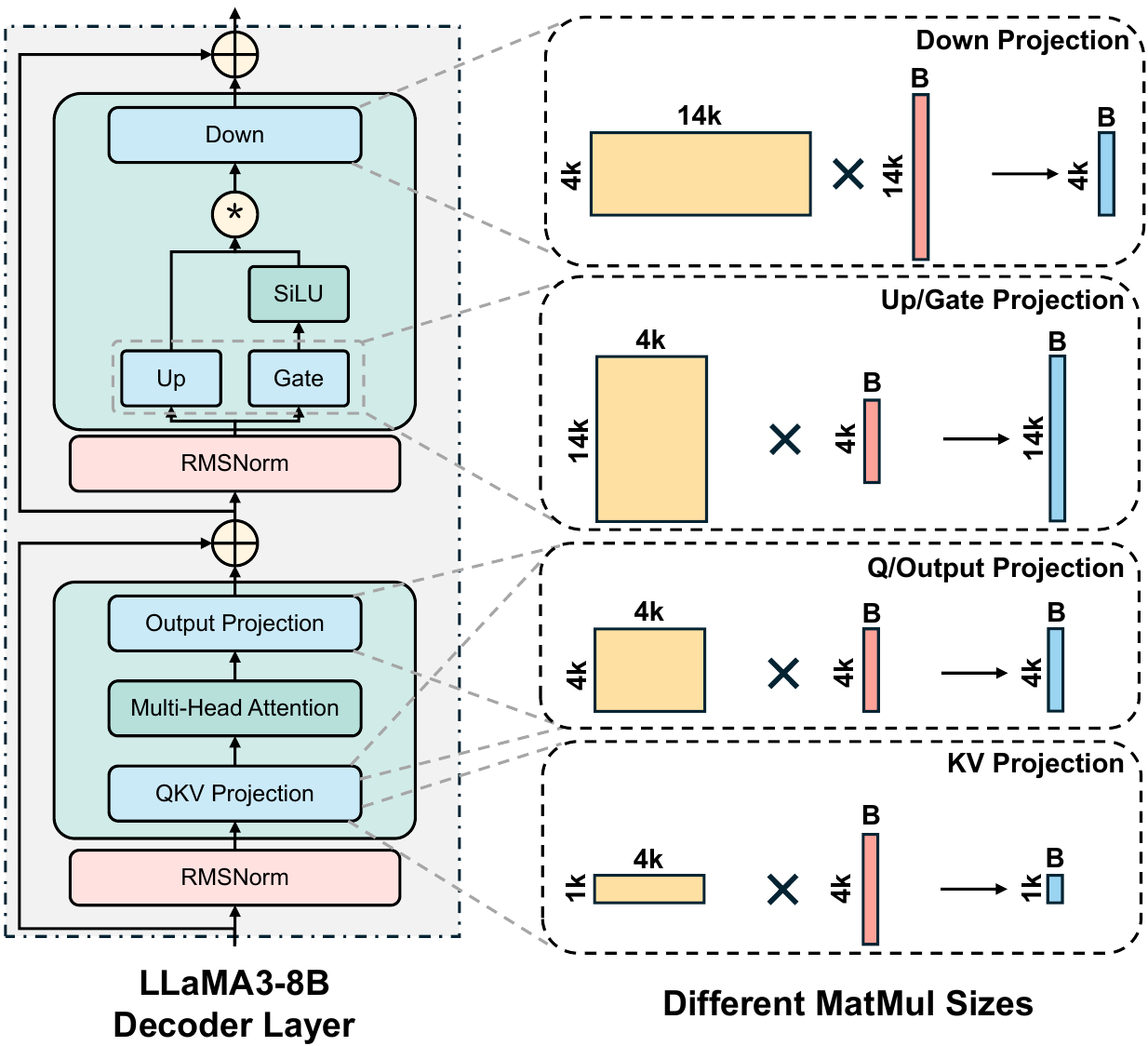}
        \vspace{+1.5em}
        \caption{The MatMul sizes vary significantly across different layers in the LLaMA3-8B model~\cite{llama-3}. \rOneMSB{Here, B refers to the batch size during the model inference phase.}}
        \label{fig:matmul different}
        \vspace{-1.5em}
    \end{figure}
\end{enumerate}

%\rOneMSB{To address the above challenges, we propose \textbf{\underline{A}rbitrary-\underline{P}recision \underline{T}ensor Core (APT)}, a GPU acceleration \rOneFC{scheme} for LLMs that supports arbitrary-precision MatMuls.} 
\rTwoMSB{To address the above challenges, we propose a GPU acceleration scheme for LLMs, namely \textbf{APT-LLM}, which features an \textbf{\underline{A}rbitrary-\underline{P}recision \underline{T}ensor Core (APT)}.}
1) \textbf{At the format level}, to address the limited precision support of GPU TCs, we propose a bit-wise decomposition and reconstruction approach for MatMuls, enabling arbitrary precision support. This method not only reduces memory overhead during LLM inference but also maintains flexibility in supporting arbitrary precision. 
Furthermore, we introduce a novel data format, bipolar-INT, which utilizes a symmetric quantization range and eliminates redundant sign bits. 
% This format is better aligned with the data distribution of LLMs and is more conducive to efficient parallel processing using Single Instruction, Multiple Data (SIMD) architectures.
\rTwoMSB{This data format can seamlessly replace the INT format without any loss of accuracy and is more conducive to efficient parallel processing using GPU architectures.}
2) \textbf{At the memory level}, we propose a memory management approach focused on efficient data recovery
% designed to maximize the use of shared memory for tasks that would otherwise be handled by global memory. 
\rOneFC{that prioritizes data recovery and maximizes shared memory utilization for tasks typically handled by global memory.}
This significantly accelerates kernel execution speed.
% 3) \textbf{At the kernel level}, to overcome the limitations of a single kernel for all matrix computations, we identified all tunable parameters within the APT kernel and established the mathematical relationships governing them. Based on these relationships, we can determine the most efficient parameter configuration for each MatMul operation by considering its size and precision, significantly improving model execution efficiency. 
3) \textbf{At the kernel level}, \rOneFC{we introduce a hyperparameter-optimization approach that overcomes the limitations of using a single kernel for all matrix computations. By identifying and formalizing the mathematical relationships between all tunable hyperparameters within the APT kernel, we establish a framework for determining optimal hyperparameter configurations. This optimization process considers both the matrix dimensions and precision requirements of each specific MatMul operation, resulting in significant improvements to model execution efficiency.}
% In experiments, we achieved a speedup of xxx $\times$ compared to xx, demonstrating the effectiveness of our approach.

% In this paper, we introduce a novel data format, bipolar-INT, which utilizes a symmetric quantization range and eliminates redundant sign bits. This format is better aligned with the data distribution of LLMs and is more conducive to efficient parallel processing using SIMD (Single Instruction, Multiple Data) architectures. 
% Furthermore, to address the limited precision support of GPU TCs, we propose a bit-wise decomposition and reconstruction approach for MatMul, enabling arbitrary precision support. This method not only reduces memory overhead during LLM inference but also maintains flexibility in supporting arbitrary precision. 
% Next, we propose a memory management approach focused on efficient data recovery, designed to maximize the use of shared memory for tasks that would otherwise be handled by global memory. This significantly accelerates kernel execution speed. Finally, to overcome the limitations of a single kernel for all matrix computations, we propose a kernel mapping method that selects the most efficient kernel based on parameters like matrix size and quantization bit-width. This ensures optimal performance for matrix inference, significantly improving model execution efficiency. In experiments, we achieved a speedup of xxx $\times$ compared to xx, demonstrating the effectiveness of our approach.

\begin{figure}[t]
    \centering
    \includegraphics[width=\linewidth]{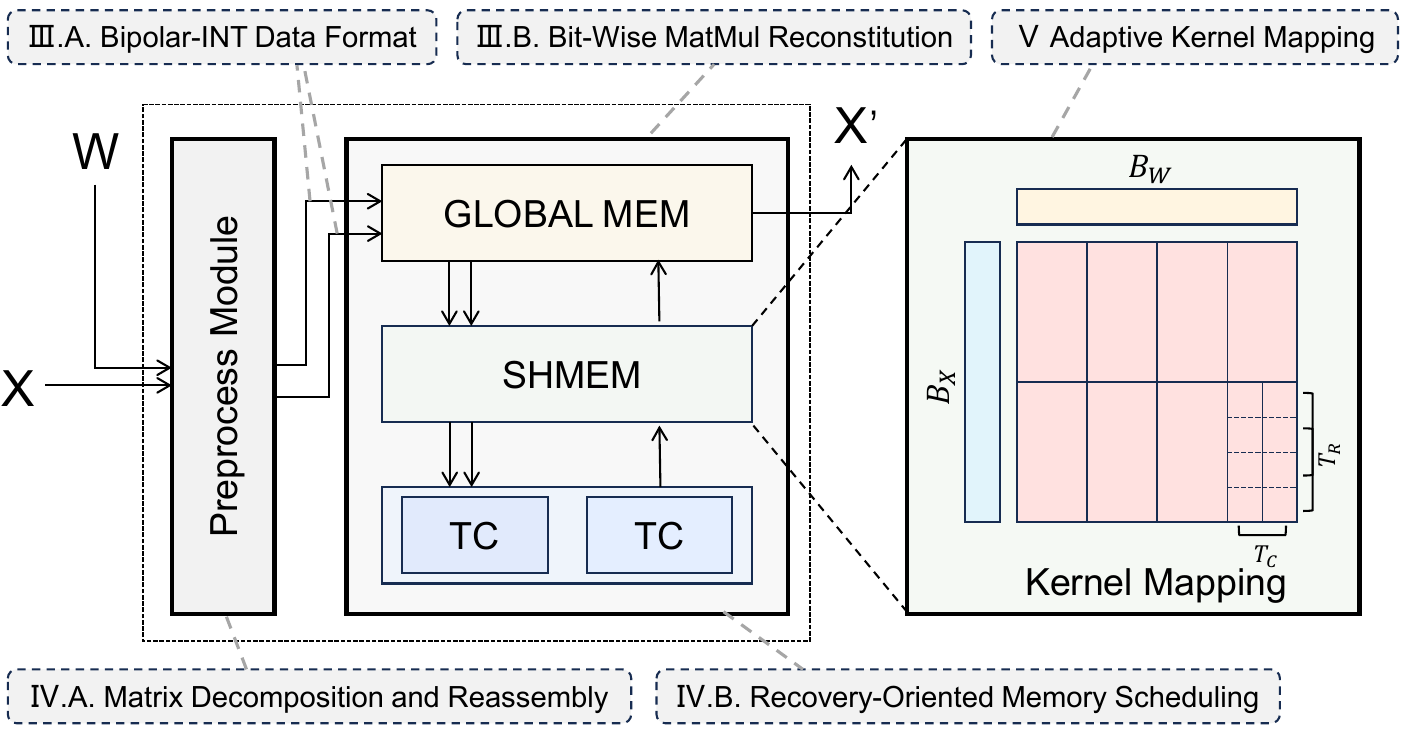}
    % \caption{\rOneFC{Overview of our optimization scheme across format, memory, and kernel levels. 
    \vspace{+0.5em}
    \caption{\rOneMSB{The proposed APT-LLM scheme provides comprehensive optimizations
    % , from algorithmic enhancements to hardware deployment
    across format, memory, and kernel levels, significantly improving the inference performance of arbitrary-precision quantized LLMs on GPUs.}}
    \label{fig:overview}
    \vspace{-1em}
\end{figure}

In summary, \rOneFC{as shown in Fig.~\ref{fig:overview}, }the main contributions of this paper are as follows:

\TODO{
\begin{itemize}
    \vspace{-1em}
    % \item We propose a novel data format called bipolar-INT for efficient arbitrary precision MatMuls, featuring a symmetric range and eliminating redundant sign bits. (Sec.~\ref{subsec:bipolar_int})
    % \item We present an innovative design for arbitrary precision MatMuls on GPUs, optimizing memory usage without sacrificing flexibility. (Sec.~\ref{subsec:bitwise_matmul})
    % \item We propose a novel MatMul design based on matrix decomposition, which supports arbitrary-precision computation while reducing both storage and computational overhead without sacrificing computational flexibility (Sec.~\ref{subsec:bitwise_matmul}). Furthermore, we introduce a new data format, bipolar-INT, which features a symmetric quantization range and eliminates redundant sign bits. This design enhances LLM quantization efficiency while improving parallel computation capabilities (Sec.~\ref{subsec:bipolar_int}). 
    \item \rTwoMSB{We introduce a new data format, bipolar-INT, which features a symmetric quantization range and eliminates redundant sign bits, enhancing LLM quantization efficiency. Simultaneously, it improves capabilities for parallel computation while preserving accuracy (Sec.~\ref{subsec:bipolar_int}). Furthermore, we propose a novel arbitrary precision MatMul design based on matrix decomposition, which supports arbitrary-precision computation while reducing both storage and computational overhead without sacrificing computational flexibility (Sec.~\ref{subsec:bitwise_matmul}).}
    % \item A hierarchical memory management strategy oriented by efficient data recovery is proposed, which greatly enhances kernel execution speed. (Sec.~\ref{sec:mem_scheduling})
    % \item We propose a novel GPU hierarchical memory management strategy designed for efficient data recovery, enabling high-performance arbitrary-precision MatMul (Sec.~\ref{subsec:memory_scheduling}). To further enhance computational efficiency, we introduce a matrix preprocessing module that leverages matrix decomposition and recomposition to eliminate redundant storage overhead while facilitating subsequent computations (Sec.~\ref{subsec:matrix_d&r}).
    \item \rOneMSB{We develop a matrix preprocessing module that leverages matrix decomposition and recomposition to eliminate redundant storage overhead while facilitating subsequent computations (Sec.~\ref{subsec:matrix_d&r}). Furthermore, to enhance data recovery efficiency and enable high-performance arbitrary-precision MatMul, we propose a GPU hierarchical memory management strategy designed for optimal memory scheduling (Sec.~\ref{subsec:memory_scheduling}).}
    % \item We introduce a kernel mapping method that dynamically selects the most efficient kernel, achieving significant performance improvements. (Sec.~\ref{sec:apt_mapping})
    \item We propose a kernel mapping method that first identifies all tunable hyperparameters within the APT kernel and establishes their underlying relationships. Based on matrix size and precision, our method dynamically selects the most efficient hyperparameter configurations, enabling high-performance model inference (Sec.~\ref{sec:apt_mapping}).
\end{itemize}
}

% \rTwoMSB{Comprehensive evaluations across several popular LLMs demonstrate that APT-LLM achieves up to a 3.99$\times$ speedup compared to FP16 baselines and a 2.16$\times$ speedup over NVIDIA CUTLASS INT4 acceleration, highlighting its promising potential for efficient LLM deployment in resource-constrained environments.}
\rThreeMSB{Comprehensive evaluations across several popular LLMs demonstrate that APT-LLM achieves up to a 3.99$\times$ speedup compared to FP16 baselines and a 2.16$\times$ speedup over NVIDIA CUTLASS INT4 acceleration on the RTX 3090. On the RTX 4090 and H800, APT-LLM achieves up to 2.44$\times$ speedup over FP16 and 1.65$\times$ speedup over CUTLASS integer baselines.}

\section{Background} \label{sec:background}   

\subsection{GPU Hierarchy and Tensor Core}
% \rOneMSB{Before delving into other aspects of the research, let's first examine the primary sources of computational power for modern LLM inference, which will help us better understand the approach to accelerating LLM inference.}
GPUs have become essential for AI workloads due to their highly parallel computing structure\cite{warp-aware,DG-replace,spmm}, which is driven by Streaming Multiprocessor (SM), the specialized cores designed to handle parallel processing tasks. 

\rOneMSB{As shown in Fig \ref{fig:overview}, }modern GPUs~\cite{a100, bstc, tc_benchmark, btc} feature a multi-level memory hierarchy, including global memory, shared memory \rOneFC{(SHMEM)}, registers, and caches (L1 and L2), each with different sizes and access speeds. Global memory, the largest and slowest, is accessible by all threads, while shared memory, though smaller, is faster and accessible within a block, reducing latency. Registers, the fastest memory component, store frequently accessed variables for individual threads. L1 and L2 caches further expedite data access, with L1 being faster and L2 larger and shared among cores\cite{ampere-benchmark,turing-benchmark,hopper-benckmark}.

% To accelerate deep learning workloads, NVIDIA introduced TCs in their GPUs.
\rOneFC{Within this memory hierarchy, NVIDIA introduces TCs to accelerate deep learning workloads\cite{gtco}.}
% Optimized for MatMuls, a critical operation in LLMs, TCs leverage 
% the massive parallelism of GPUs 
% \rOneFC{GPU parallelism}
% to significantly improve computational efficiency and inference performance
\rOneMSB{Optimized for MatMuls, a critical operation in LLMs, its operation is based on the SIMT (Single Instruction, Multiple Threads) architecture, which relies on the Warp mechanism in GPUs. A Warp consists of 32 threads that execute the same instruction in parallel. 
Each time a Tensor Core performs a computation, it can be divided into several fragments, which are the smallest data units processed by the Tensor Core. In GPU computing,  fragments represent the way data is partitioned for processing.
Through efficient thread scheduling and resource allocation, Tensor Cores are able to execute MatMuls and other high-throughput operations with improved computational efficiency\cite{ampere-benchmark,turing-benchmark,hopper-benckmark,TC-SHMEM,TC-GNN,tc_benchmark}.}

\subsection{Prefill vs. Decode in LLM Inference}
\begin{figure}[t]
	\centering
	\includegraphics[width=0.9\linewidth]{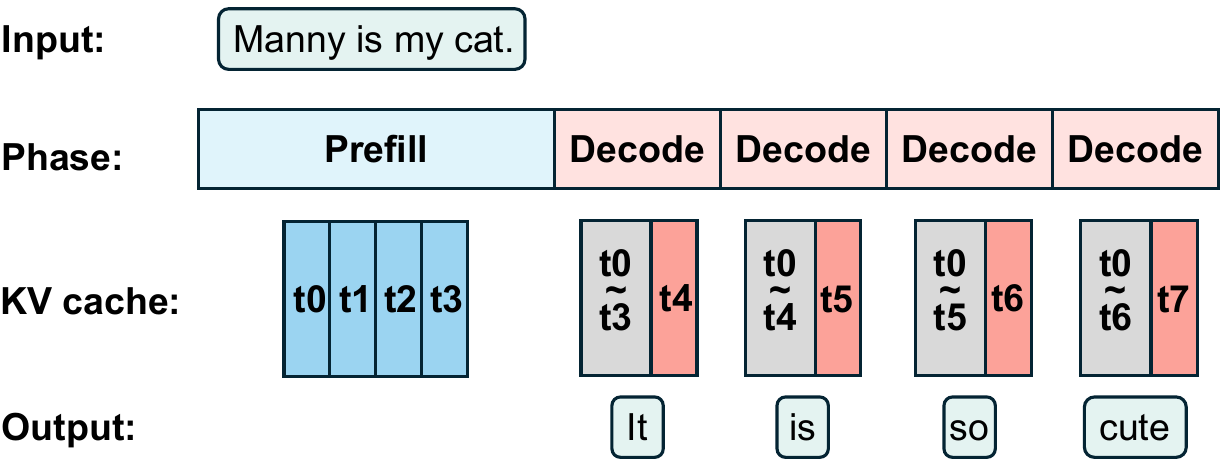}
        \vspace{1em}
	\caption{Comparison between the prefill and decode phases of LLM inference. The prefill phase involves processing multiple tokens in parallel, while the decode phase generates tokens sequentially, with each token depending on the previous ones for context.}
	\label{fig:phases-in-llm}
        \vspace{-1em}
\end{figure}

% \rOneMSB{Having understood the GPU architecture, the next step is to explore how LLMs perform inference on GPUs.}
\rOneFC{Building on the overview of GPU architectures, we describe how LLM inference executes on GPUs.}
As shown in Fig.~\ref{fig:phases-in-llm}, the process of LLM inference consists of two separate stages: the prefill phase and the decode phase, each exhibiting distinct computational properties\cite{survey}.

\rOneMSB{In the prefill phase, the model processes the entire input sequence in parallel, calculating the hidden states for all tokens simultaneously. This phase is computationally intensive, primarily involving general matrix-matrix multiplication (GEMM) operations that take full advantage of the parallel processing capabilities and efficiency of TCs in GPUs~\cite{prefill}. Additionally, key-value (KV) caching is employed to store intermediate key and value activations, which are reused during the decode phase to avoid redundant calculations and improve efficiency. This caching mechanism helps manage memory usage and accelerates inference by allowing the model to retrieve previously computed information for each new token~\cite{kvcache,kvcache-1}.} While this parallelism optimizes throughput, the substantial memory requirements \rOneMSB{for storing KV cache} can become a bottleneck, especially when handling longer input sequences\cite{kvcache}.

In the decode phase, the model generates tokens sequentially in an autoregressive manner, where each new token depends on the previously generated ones. Unlike the prefill phase, parallelism is limited because only one token is generated per step. As a result, many GEMM operations are reduced to general matrix-vector multiplication (GEMV) tasks, which are less optimized for TCs. The efficiency of GEMV computation and memory access becomes a critical factor in determining both the performance and power consumption of LLM inference during this phase~\cite{llmviewer}.

\subsection{Ultra-Low-Bit Quantized Models}
The rapid development of LLMs has led to a significant increase in model size and computational complexity, posing challenges for efficient inference. 
% To address this issue, model quantization has emerged as a promising approach, aiming to reduce the computational complexity and memory footprint of LLMs. 
% \rOneFC{Lower bit-width directly correlates to reduced memory usage and increased computational efficiency, with each reduction from FP32 to FP16, INT8, and beyond offering progressive improvements in throughput and resource utilization.}
\rOneFC{Model quantization\cite{llmint8, qlora, gptq, squeezellm, tsld, onebit} has emerged as a promising approach, which progressively improves efficiency as quantization levels decrease from FP32 to FP16, INT8, and beyond, optimizing both memory usage and computational throughput.}

Early LLM quantization methods primarily focused on traditional techniques such as FP16 optimization and INT8 quantization. For example, GPT3.INT8()\cite{llmint8} tackled the issue of outliers in quantized LLMs by employing mixed-precision computation. However, to push the boundaries further, more aggressive quantization methods with lower bit-widths have been introduced. These include QLoRA with 4-bit\cite{qlora}, GPTQ with 3-4 bits\cite{gptq}, SqueezeLLM with 3-bit\cite{squeezellm}, TSLD\cite{tsld} with ternary quantization, and OneBit\cite{onebit} with binary quantization for LLMs.

\rThreeMSB{Moreover, recent advances, such as QuantSR\cite{qin2023quantsr} and IR-QLoRA\cite{qin2024accurate}, demonstrate high performance with low-bit quantization for both super-resolution and LLMs, while BiMatting\cite{qin2023bimatting}, BiBERT\cite{qin2022bibert} and BeBERT\cite{bebert} achieve remarkable efficiency through extreme binarization. These works underscore the inherent trade-off between reduced numerical precision and the resulting computational efficiency gains in both vision and language tasks.}

Despite the promising results achieved by these ultra-low-bit quantization methods, their optimal inference performance on GPUs has been hindered by the lack of suitable data formats supported by GPU hardware. This limitation calls for efficient acceleration designs that can bridge the gap between the quantization methods and the available GPU hardware capabilities, enabling LLMs to fully benefit from the reduced precision and computational complexity offered by ultra-low bit quantization.

\subsection{Arbitrary Precision Acceleration Schemes}
\rOneMSB{To enable hardware support for arbitrary-precision model inference,} arbitrary precision acceleration schemes for data formats lower than INT8 have been extensively studied\cite{anda,courbariaux2015binaryconnect,apnn,o3bnn,o3bnn-r,deep-compression,bstc,btc,Turing,olacc,wang2019haq,lq-nets,dorefa,bebert} to optimize inference performance and maintain model accuracy for diverse application requirements. 
In addition to the commonly supported precisions on modern GPUs (e.g., INT1, INT4), these designs often incorporate a broader range of precisions such as INT2 and INT3. 

\rOneMSB{Furthermore, mixed-precision quantization introduces additional challenges for GPU inference, with settings like W4A8 (4-bit weights, 8-bit activations)\cite{qserve}, W1A8\cite{Wang2023BitNetS1}, and W1A16\cite{onebit} complicating efficient execution. GPUs and TCs lack support for efficient arbitrary precision operations. This limitation hinders the efficient acceleration of ultra-low-bit mixed-precision quantized LLMs, necessitating a novel acceleration scheme that fully utilizes TCs to perform quantized LLMs with arbitrary precision\cite{apnn}.}

Among these schemes, APNN-TC\cite{apnn} supports arbitrary precision, HAQ\cite{wang2019haq} employs 1-8 bits, while BSTC~\cite{bstc} and BTC~\cite{btc} focus on 1-bit precision. These works designed MatMul kernels supporting various precisions and integrated them into quantized neural networks to enhance performance. However, they have limitations in terms of incomplete support for arbitrary precision, as exemplified by APNN-TC not supporting W3A4, and suboptimal performance for large matrix parameters due to unsuitable data formats and inefficient GPU memory management. To address these issues, our work aims to identify a more suitable data format for arbitrary precision MatMul, implement kernel support for true arbitrary precision, and further optimize MatMul performance through improved memory scheduling.
% 【1页】
% \begin{itemize}
%     \item 超低比特量化LLM介绍: 超低比特量化工作介绍
%     \item GPU 架构介绍: 多级存储架构以及Tensor Core基本信息, 体现其多级存储, 支持少量数据格式等特点
%     \item 已有任意精度支持工作介绍: BSTC, APNN, ABQ等
% \end{itemize}
\section{Arbitrary Precision MatMul}\label{sec:ap_matmul}
% \rOneMSB{First, we need to consider how to efficiently implement arbitrary-precision matrix multiplication on GPUs from the perspectives of both data format and computation methods.}
% This section outlines the framework for arbitrary precision integer MatMul implemented on TCs\rOneMSB{, where integer-based formats, particularly fixed-point integers, are the primary data format used in current quantization efforts.}. 
% Initially, we introduce an efficient data format, referred to as bipolar-INT, highlighting its benefits over conventional signed and unsigned integers when deploying on TCs. Subsequently, we propose a bitwise MatMul decomposition technique, which dissects the operands at the bit level using our bipolar-INT format and leverages the bitwise computing kernel capabilities of TCs, followed by the development of a data recovery dataflow.

\rOneFC{This section presents a framework for arbitrary precision integer MatMul on TCs, tackling the growing need for diverse ultra-low-bit quantization methods in LLMs.}
\rOneFC{Our approach centers on fixed-point INTs that is the dominant data types in quantization techniques.}
\rOneFC{We first introduce bipolar-INT, an efficient data representation that overcomes limitations of conventional signed INT when deployed on TCs.}
\rOneFC{Building on this format, we develop a bitwise MatMul decomposition method that breaks down operands at the bit level and leverages TC bitwise computing capabilities, followed by our proposed data recovery dataflow.}

\subsection{Bipolar-INT Data Format} \label{subsec:bipolar_int}
We present an innovative and efficient data format named bipolar-INT, designed for arbitrary precision MatMul computations. In contrast to traditional signed INT, bipolar-INT is better suited for quantization in LLMs and parallel computation because of its 
%symmetric range and 
standardized operations\rOneFC{, making it more efficient when implementing on TCs.} Moreover, we propose a hardware-efficient and numerically lossless conversion scheme from signed INT to bipolar-INT.
% . This makes it especially beneficial for implementation on TCs.

\begin{figure}[t]
    \centering
    \includegraphics[width=0.85\linewidth]{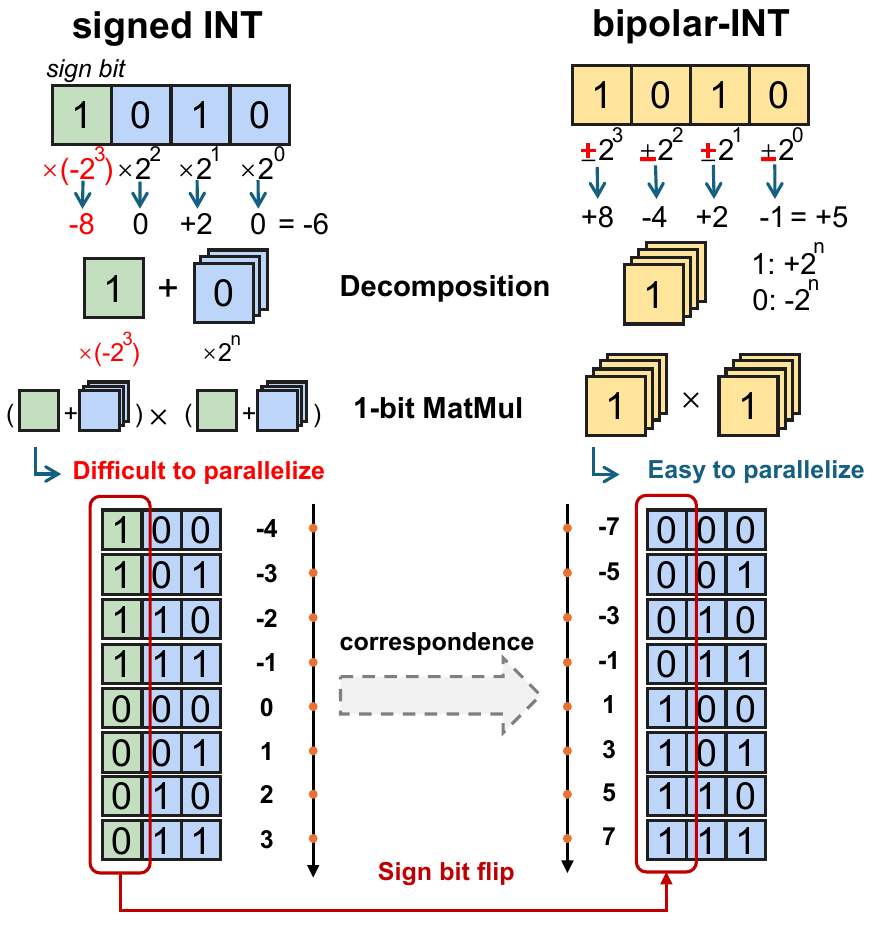}
    % \caption{Compared to original signed and unsigned integers, bipolar-INT is suitable for parallel computing due to symmetric quantization and unified operations.}
    \vspace{1.5em}
    \caption{\rTwoMSB{Comparison between bipolar-INT and signed INT. Bipolar-INT is well-suited for TCs’ parallel computing due to its unified operations and can be easily derived from INT.}}
    \label{fig:bipolar compare}
    \vspace{-1.5em}
\end{figure}

As shown in Fig. \ref{fig:bipolar compare}, the key difference between bipolar-INT and signed INT lies in the interpretation of each bit.
% In traditional integers, each bit except the sign bit is valued as 0 or 1, whereas in bipolar-INT, the "0" is interpreted as "-1" in calculations, allowing each bit to be either -1 or 1. 
In signed INT representation, each bit represents a value of 0 or 1 aside from the sign bit, whereas bipolar-INT reinterprets the 0 value as -1, allowing each bit to represent either -1 or 1 during computation.
Specifically, for an $n$-bit bipolar-INT data $x = x^{(n-1)}, . . . , x^{(1)}, x^{(0)}$, its decimal value can be obtained by
$$(x)_D = \sum\nolimits_{i=0}^{n-1}(2x^{(i)}-1)\cdot 2^i,$$
\rOneFC{where $i$ denotes the bit position of $x$.}

% From the perspective of data distribution, the weights of LLMs typically exhibit a symmetric normal distribution\cite{zeng2024abq}. However, standard quantization formats, whether unsigned INT or signed INT, disrupt this symmetry. For example, a 2-bit quantization range for unsigned INT is (0, 1, 2, 3), while for signed INT it is (-2, -1, 0, 1). Both formats deviate from the original symmetric weight distribution. In contrast, the bipolar-INT format inherently features a symmetric quantization range, which better aligns with the natural symmetry of LLM weights and is therefore more suitable for weight quantization in LLMs.

% \rTwoMSB{From the perspective of data distribution, LLM data usually have a zero-centered normal distribution \cite{qlora}, which exhibits a symmetric normal distribution and the majority of data is distributed near zero. However, when utilizing unsigned INTs for quantization, all data is quantized to non-negative integers. For example, when a model is quantized to 8 bits, data values close to zero are quantized to integers greater than 100. This significantly increases the risk of data overflow during MatMul operations.}

From the perspective of parallel computation, in the context of signed INT quantization, as shown in Fig. \ref{fig:bipolar compare}, the use of two’s complement arithmetic causes the 
% MSB 
\rOneFC{most significant bit}
matrix to have a sign opposite to that of the remaining bits after decomposition. 
This discrepancy necessitates specialized handling in both MatMul operations and matrix reconstruction. 
Such a requirement introduces substantial challenges to the parallel computation of single-bit matrices at each bit level.

\rTwoMSB{Moreover, signed INT can be precisely converted to bipolar-INT without affecting the accuracy of quantized models. 
Linear quantization is fundamentally represented by the equation:}
\rTwoMSB{$$W=s\hat W+z,$$
where $W$ and $\hat W$ denote the matrices before and after quantization, respectively. The parameters $s$ (scale) and $z$ (zero) are two critical factors that control the quantization granularity and zero-point offset value in linear quantization, respectively. As shown in Fig. \ref{fig:bipolar compare}, in binary terms, simply flipping the sign bit of a signed INT yields the corresponding bipolar-INT data. In numerical terms, correspondence between bipolar-INT and signed INT is:}
\rTwoMSB{$$\hat x'=2\hat x+1,$$
where $\hat x'$ and $\hat x$ denote bipolar-INT data and INT data. Substituting this relationship into the linear quantization formula yields:}
\rTwoMSB{$$W=s\frac{\hat W'-1}{2}+z,$$
$$W=\frac{s}{2}\hat W'+(z-\frac{s}{2}),$$
where $\hat W'$ is the matrix after bipolar-INT quantization. Thus, by regarding $\frac{s}{2}$ as the new scale and $z-\frac{s}{2}$ as the new zero, we can obtain bipolar-INT quantization parameters that achieve the exact same precision as INT quantization. Moreover, the scale and zero are stored in a high-precision data format, which ensures that there is no reduction in precision.}

%Moreover, for binary quantized neural networks\cite{onebit,Wang2023BitNetS1}, weights ($W$) are often quantized to values of either -1 and 1, which are encoded as 1-bit 0 and 1, respectively. If the feature matrix $X$ is quantized to (0,1) with its decimal value also being (0,1), then the MatMul of $W$ and $X$ will be inconsistent during computation. Conventional methods like APNN-TC\cite{apnn} introduce an additional all-ones matrix $J=[1,1]$ to tackle this problem. The matrix $W$ is decoded as $\hat{W} = [0, 1]$, which relates to its actual value by $W = 2\hat{W}-J$. Thus, the MatMul becomes $W X = 2\hat{W}X-JX$, which not only introduces an additional matrix $J$ occupying memory, but also introduces an extra MatMul operation $JX$. In contrast, 
% the binary quantized $W$ aligns perfectly with our 1-bit bipolar-INT format. 
%\rOneMSB{as shown in Fig. \ref{fig:bipolar compare}, when using 1-bit precision, the bipolar-INT quantization method represents -1 and 1 with 0 and 1, perfectly aligning with the binary quantization of weights. }Therefore, when using bipolarINT for MatMuls in binary quantized neural networks, the feature matrix $X$ is also quantized using bipolar-INT, without introducing additional matrices or MatMuls for precise computation.

\subsection{Bit-Wise MatMul Reconstitution}\label{subsec:bitwise_matmul}

\begin{figure}[t]
    \centering
    \includegraphics[width=0.90\linewidth]{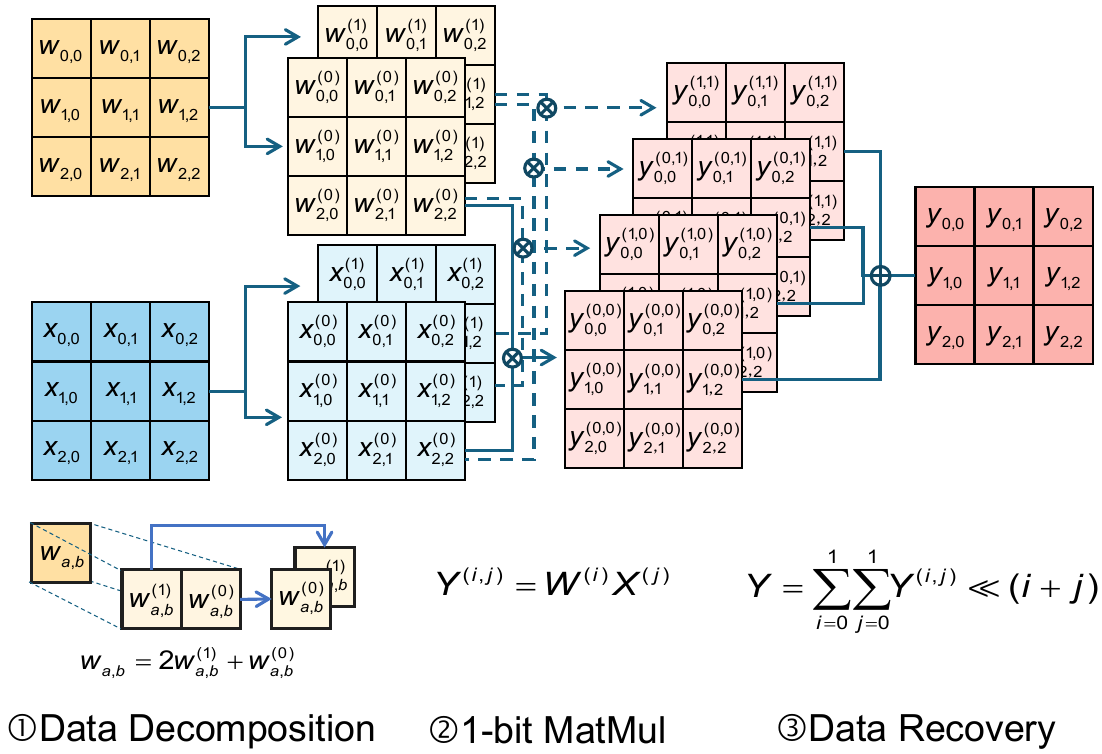}
    % \caption{Illustration of the computation process of arbitrary precision MatMul. Here matrices W and X are both 2-bit, which can be extended to arbitrary bit widths.}
    \vspace{1.5em}
    \caption{Illustration of the \rOneMSB{Bit-wise MatMul Reconstitution process} for bipolar-INT MatMuls. Here matrices W and X are both 2-bit, which can be extended to arbitrary bit widths.}
    \label{fig:AP-bit-matmul}
    \vspace{-1.5em}
\end{figure}

\rOneMSB{To address the issue of TCs not supporting arbitrary precision computation, w}e introduce a bit-wise MatMul reconstitution approach grounded in our bipolar-INT format. This approach encompasses three fundamental stages: data decomposition, 1-bit MatMul, and data recovery. Fig.~\ref{fig:AP-bit-matmul} demonstrates the entire computational process for achieving arbitrary precision MatMul utilizing our proposed method. In this illustrative example, we examine a MatMul operation that includes 2-bit matrices $W$ and $X$, culminating in the computation of a 32-bit output $Y = WX$.

The process of data decomposition entails breaking down the operands on a bit-by-bit basis, thereby facilitating the application of the bit-wise computing kernel that TCs provide. 
% In particular, both $W$ and $X$ are partitioned into two distinct matrices each, referred to as $W^{(i)}$ and $X^{(j)}$. In this context, $i$ and $j$ represent the position of the 1-bit matrix element within the original matrix \rOneFC{of $W$ and $X$, respectively.}
\rOneFC{In particular, both $W$ and $X$ are partitioned into bit-level matrices $W^{(i)}$ and $X^{(j)}$, where indices $i$ and $j$ represent the position of each bit.}
% as specified in Equ.~\ref{equal:1}.}
% corresponding to the $i$ in Equ.~\ref{equal:1}.
Subsequent to the decomposition of data, pairwise 1-bit MatMul operations are executed on the resulting operands. NVIDIA GPUs facilitate the utilization of either AND or XOR logic gates for executing 1-bit MatMul operations within TCs. The pairwise multiplication of $W^{(i)}$ and $X^{(j)}$ yields a 32-bit intermediate result matrix $Y^{(i,j)}$ for each corresponding bit pair.
In order to derive the final outcome from the intermediate bit-wise calculations, a data recovery dataflow is utilized. This process entails reconstructing the output matrix $Y$ from the intermediate matrices $Y^{(i,j)}$. Each $Y^{(i,j)}$ is shifted according to its respective bit positions $(i,j)$, after which all the shifted matrices are aggregated through summation.

While the example depicted in Fig.~\ref{fig:AP-bit-matmul} utilizes 2-bit matrices, the underlying principles are applicable to matrices of any bitwidth. Our decomposition approach for bit-wise matrix multiplication, leveraging the bipolar-INT format, capitalizes on the bit-wise computational capabilities of TCs, thereby offering a versatile and efficient method for performing matrix multiplication operations with arbitrary precision.
% \begin{itemize}
%     \item 比特级拆分的方式, 对1bit矩阵进行矩阵乘法运算, 在运算后移位相加
%     \msb{\item[+] 比特级重组的计算方式与一般INT计算数学等效的证明}
% \end{itemize}
\section{GPU Memory Scheduling}\label{sec:mem_scheduling}
% \rOneMSB{Specifically, to implement the aforementioned method on GPUs, we need to optimize the GPU deployment strategy in terms of both data transfer and memory management.}
\rOneFC{While our bit-level decomposition method establishes the computing approach for arbitrary precision MatMul, efficient GPU implementation requires careful memory optimization.}
This section examines strategies for optimizing data transfer across various levels of the memory hierarchy within the arbitrary precision MatMul kernel to enhance processing speed.

\subsection{Matrix Decomposition and Reassembly}\label{subsec:matrix_d&r}
\rOneMSB{To enhance data transfer efficiency, we propose a method for decomposing and reassembling matrices across different levels of the memory hierarchy within the arbitrary precision MatMul kernel.} This approach is designed to minimize redundant memory access and enhance the speed of data transfer by preprocessing the initial $n$-bit INT matrix.

\begin{figure}[tb]
    \centering
    \includegraphics[width=0.90\linewidth]{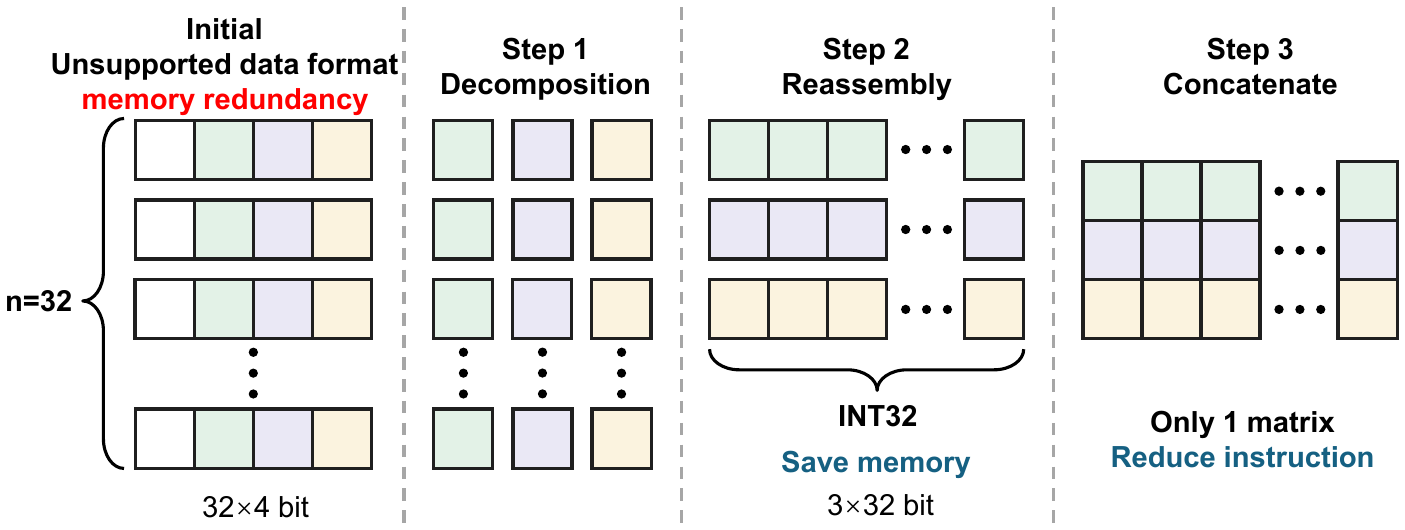}
    % \caption{Steps for matrix decomposition and reassembly to save memory and reduce instruction.}
    \vspace{1.5em}
    \caption{Procedure of matrix decomposition and reassembly to \rOneFC{reduce GPU memory consumption and number of data transfer instructions.}}
    \label{fig:decomposition and reassembly}
    \vspace{-1.5em}
\end{figure}

GPUs accommodate a wider range of precisions than TCs in terms of memory access, yet they do not encompass every conceivable precision. For instance, a 3-bit INT type lacks an appropriate storage format, necessitating storage in a broader format (such as 4-bit or 8-bit), which results in additional memory access overhead. Furthermore, when accessing substantial quantities of data, it is crucial to consider not only the support for storage precision but also the compatibility of the associated data format for data transfer.

To tackle these challenges, our approach comprises three steps, depicted in Fig. \ref{fig:decomposition and reassembly}. In Step 1, we execute a 1-bit decomposition of the original matrix, disassembling each bit and reassembling them with corresponding bits from other data sources to create $n$ 1-bit matrices. This step addresses the problem of incompatible data formats and eradicates memory redundancy due to the absence of suitable data formats.
In Step 2, we proceed by reconstituting the decomposed data with 32-bit unsigned integers. This procedure guarantees compatibility with the GPU’s native support, thus \rOneMSB{there’s no need for complex memory mapping or copying, and data can be directly placed into the GPU’s memory, reducing the latency associated with intermediate steps.}  Subsequently, Step 3 involves methodically concatenating the processed $n$ matrices into one unified matrix. This approach not only optimizes memory usage but also simplifies $n$ individual data transfer operations into a single command. Despite the unchanged volume of transferred data, this concatenation enhances the transfer speed and preserves storage capacity.
Following this sequence of procedures, our method for matrix decomposition and reconstruction substantially reduces unnecessary memory access and enhances data transfer rates within the arbitrary precision MatMul kernel.
% \begin{itemize}
%     \item 输入矩阵逐比特拆解, 并按照相同比特位重组, 以便于后续进行MatMul运算
%     \msb{\item[+] 预处理模块(GPU中高效实现矩阵拆解重组)的设计 }
% \end{itemize}

\subsection{Recovery-Oriented Memory Scheduling}\label{subsec:memory_scheduling}
\rOneMSB{To further enhance data transfer efficiency and memory access, this study proposes a recovery-oriented memory scheduling strategy within the arbitrary precision MatMul kernel on GPUs} by performing the matrix recovery process in SHMEM or fragments, thereby reducing global memory access and computation, and further accelerating the kernel's computation speed.

As shown in Fig. \ref{fig:AP-bit-matmul}, when implementing the arbitrary precision MatMul described in Sec. \ref{subsec:bitwise_matmul} on a GPU, we perform 1-bit MatMul on the decomposed matrices to obtain intermediate result matrices. These intermediate result matrices are then shifted and summed to get the final result. In a naive approach, each SM directly multiplies a pair of 1-bit matrices decomposed from the weights and features, and the MatMul result is directly returned to global memory for recovery. However, this design leads to each SM obtaining at most one intermediate result matrix, forcing the final matrix recovery to be performed in the slower global memory. This step introduces significant delays because accessing and processing data in global memory is much slower compared to SHMEM.

\begin{figure}[tb]
    \centering
    \includegraphics[width=1\linewidth]{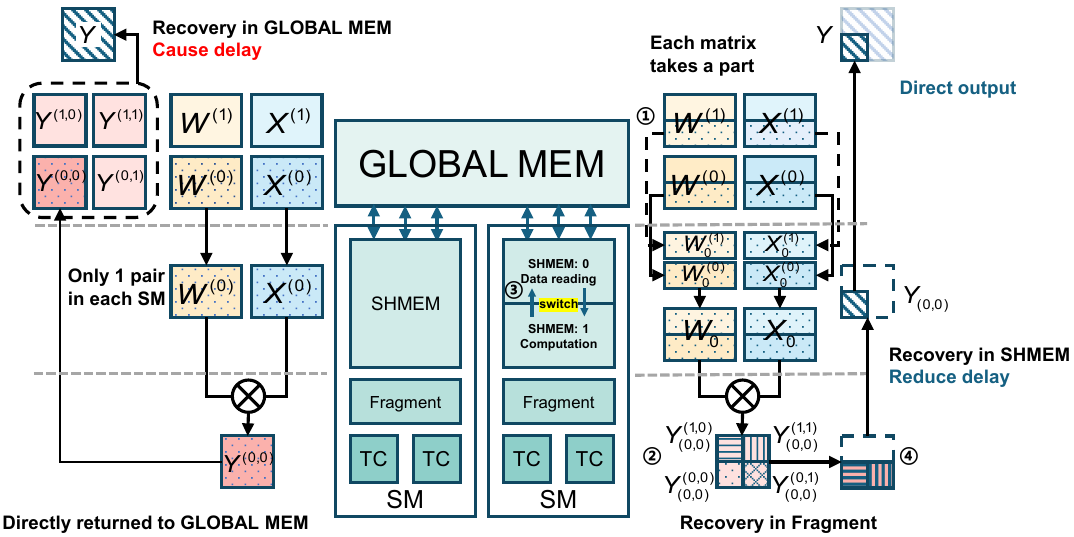}
    \vspace{1em}
    % \caption{Complete GPU multi-level memory scheduling design. Oriented by efficient data recovery, MatMul's latency is reduced.}
    \caption{Recovery-oriented memory scheduling strategy for arbitrary precision MatMul on GPUs, leveraging SHMEM and fragments to reduce global memory access and accelerate computation.}
    \label{fig:GPU-Mem-Scheduling}
    \vspace{-1em}
\end{figure}

To address this issue, our recovery-oriented memory scheduling strategy aims to compute all intermediate result matrices within a single SM, as shown in \textcircled{\scriptsize{1}} of Fig. \ref{fig:GPU-Mem-Scheduling}. This requires each SM to compute all bitwise combinations of the weight and feature matrices. To efficiently manage SHMEM, given its limited size, we divide the output matrix into blocks of size $B_M\times B_N$ \rOneMSB{($B_M$ and $B_N$ represent the number of input matrix rows and weight matrix columns processed per block)}, with each SM responsible for computing the data within one block. If the number of blocks exceeds the number of SMs, the SMs are iteratively called to perform the computations.

Due to the insufficient size of SHMEM, the dimension $K$ needs to be partitioned. \rOneMSB{Assuming the bit width of the input and weight is $p$ and $q$, and $B_K$ is the hidden dimension processed by a block in one iteration. }Each time, the SM only reads data from two matrices of size 
% $n_{w,x}b_{m,n}\times b_k$
$pB_M\times B_K$ and $qB_N\times B_K$, and the results of each computation are accumulated over $K/b_k$ iterations to produce the complete output value. In SHMEM, the input weight and feature matrices of different bits are concatenated into two matrices and input into the Fragment to call the TC to perform 1-bit matrix multiplication. The resulting $pB_M\times qB_N$ matrix contains all the data needed to recover a $B_M\times B_N$ output block, as shown in \textcircled{\scriptsize{2}} of Fig.~\ref{fig:GPU-Mem-Scheduling}. By sending these data back to SHMEM for data recovery, we can obtain part of the final output directly, without involving global memory in the computation.

To hide the data transfer latency from global memory to SHMEM, we allocate two blocks of SHMEM of the same size, as shown in \textcircled{\scriptsize{3}} of Fig. \ref{fig:GPU-Mem-Scheduling}. While one block is responsible for computation, the other block reads the next set of data. They then alternate in this manner, effectively overlapping data transfer and computation.
Furthermore, to increase data reuse and reduce latency, we allow each Fragment to read the weight matrix of the same bit and all bits of the feature matrix, calculating all intermediate results corresponding to this bit of the weight matrix, as shown in \textcircled{\scriptsize{4}} of Fig. \ref{fig:GPU-Mem-Scheduling}. In this way, we can perform the feature part of the data recovery in the Fragment, leaving the weight part of the recovery for SHMEM computation.

% By employing this recovery-oriented memory scheduling strategy, we significantly reduce the memory access and computation in global memory, effectively leveraging the faster SHMEM and fragments to accelerate the arbitrary precision MatMul kernel on GPUs.
% \begin{itemize}
%     \item 对比一般正常实现方式, 点出其瓶颈在于矩阵恢复需要在GLOBAL MEM中进行, 造成了巨大的延迟
%     \item 以高效恢复为导向, 想要在SHMEM中进行矩阵恢复, 需要在一个block中读取每个比特矩阵的一部分
%     \item 结合图片介绍内存调度方式
% \end{itemize}
\section{Adaptive Kernel Mapping} \label{sec:apt_mapping}

% \rOneFC{The section...}
\rOneMSB{As demonstrated in Sec.~\ref{sec:intro}, the sizes of MatMul operations vary significantly across different layers \rOneMSB{and different phases (prefill or decode)} within the same LLM. In this section, we introduce a kernel mapping method that adaptively chooses suitable kernel configurations. These configurations pertain to the specific hyperparameters and settings that determine the execution of a MatMul task on the GPU across various matrix sizes, thereby ensuring optimal performance acceleration.}

% \subsection{Overview of APT}
\subsection{Tunable Hyperparameters in APT}

% \rOneSHK{In previous sections, we have introduced the novel data format termed bipolar-INT and proposed an arbitrary-precision MatMul method based on bit-level decomposition using bipolar-INT.}
% Next, we describe the data preprocessing steps required to implement this method. Furthermore, to optimize the execution of this approach, we designed a GPU memory scheduling strategy focused on efficient data recovery. 

To further enhance the performance and flexibility of APT, we now provide a detailed overview of the computational flow of the proposed method in this work.
\rOneMSB{At the same time, we identify variable hyperparameters within the kernel configurations that can be customized to optimize the execution of MatMul for different matrix sizes, ensuring efficient use of GPU resources.} 

First, based on the GPU’s performance configuration and the desired SHMEM size, we determine the total bit width of the weights and features computed within each block, denoted as \rOneMSB{$B_w$}
and \rOneMSB{$B_x$}, respectively.
Using the bit widths of the weights and features, we then calculate the matrix sizes for each block, i.e., determining \rOneMSB{\textbf{$B_M$}} and \rOneMSB{\textbf{$B_N$}}. Compared to the common approach of first determining \textbf{\rOneMSB{$B_M$}} and \textbf{\rOneMSB{$B_N$}} and then allocating SHMEM based on bit width, our method of determining SHMEM size first and then calculating \textbf{\rOneMSB{$B_M$}} and \textbf{\rOneMSB{$B_N$}} offers several advantages. This approach ensures a stable and appropriate allocation of SHMEM, avoiding memory underutilization or overflow. Additionally, by determining the SHMEM size in advance, we can avoid the latency caused by dynamic memory allocation after computation.

\begin{figure}[tb]
  \centering
  \includegraphics[width=0.9\linewidth]{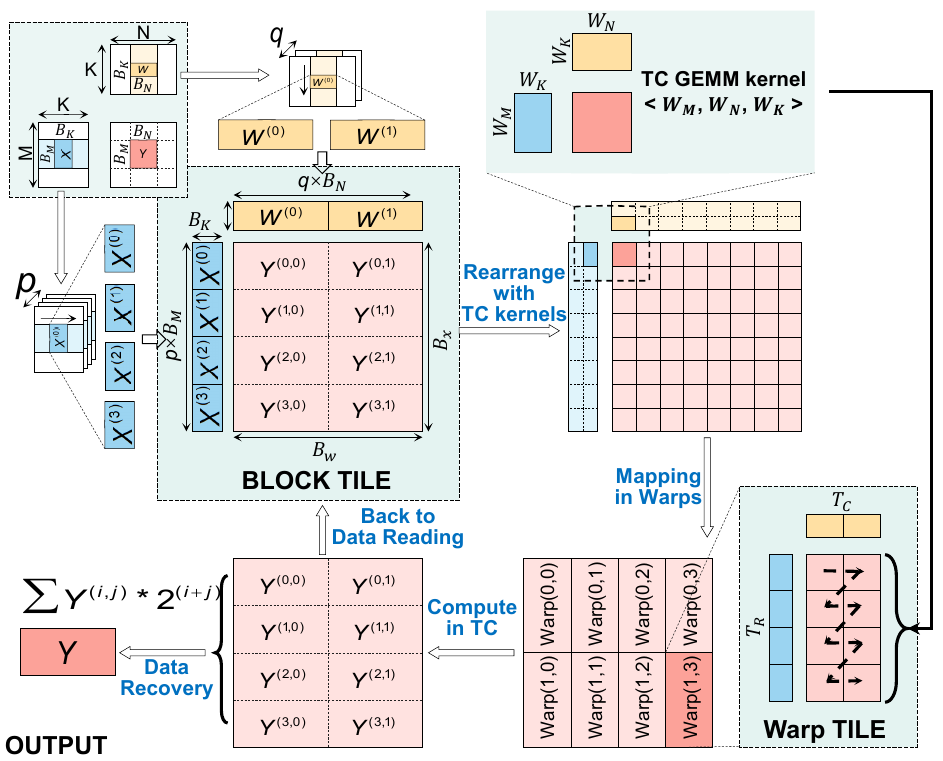}
  \vspace{1.5em}
  \caption{Detailed overview of the computational flow of APT.}
  \label{fig:overview-APT}
  \vspace{-1.5em}
\end{figure}
% 给一张大图描述本文思路在GPU中的具体实现（类似ABQ-LLM中的Fig 4），并解释其运算流程以及其中的可变参量

Subsequently, as illustrated in Fig. \ref{fig:overview-APT}, the task of multiplying an $M\times N\times K$ matrix is segmented into multiple blocks, where each block is tasked with generating a \rOneMSB{$B_M\times B_N$} output matrix. Within each block, the operation is further broken down into $\frac{K}{B_K}$ \textbf{block tile}s, with each block tile responsible for calculating a sub-MatMul of dimensions \rOneMSB{$B_M\times B_N\times B_K$}. In each block tile, using the \textbf{Bit-Wise MatMul Reconstitution} approach, the weight and input matrices are decomposed and recombined bit by bit, then stored in SHMEM with a size of $\frac{wB_K}{B_x}$. In this way, a block tile can simultaneously compute all the intermediate matrices $Y^{(i,j)}$ shown in Fig. \ref{fig:overview-APT}.

Once the data is loaded into SHMEM, we temporarily no longer need to consider the differences between the individual bits and instead treat the data as a whole for MatMul. Instead, we treat the data as a whole for MatMul. At this stage, we divide the entire block tile into several fragments and use NVIDIA's built-in 1-bit GEMM kernel to leverage TC acceleration for MatMul. For example, when invoking the 1-bit GEMM kernel with hyperparameters $\langle W_M, W_N, W_K\rangle = \langle8, 8, 128\rangle$, the input weight and feature matrices are divided into several $8\times 128$ 1-bit fragments. After MatMul, the result is an $8\times 8$ matrix in 32-bit precision.

\rOneSHK{In practice, the TCs in NVIDIA GPUs are invoked by warps within the SMs.
However, a single SM does not have enough warps to dedicate one warp to each 1-bit GEMM kernel.} 
Additionally, too many warp calls may reduce data reuse efficiency. Therefore, we need to further divide the GEMM kernel into multiple warp tiles, allowing each warp to compute multiple kernels. During the computation within each warp tile, we perform data reuse on the warp row tile. Specifically, after loading the weight data for one tile, we compute all the feature tiles for that row before loading the next weight tile.

Finally, after completing the matrix operations, we use the initially calculated \rOneMSB{$B_M$} and \rOneMSB{$B_N$} to determine which intermediate result matrix each data element belongs to within the block tile. The data is then shifted, summed, and directly written to global memory.

% 根据该流程分析GEMM和GEMV为什么不能用同一个配置，不同大小GEMV为什么不能用同一个配置，以及人工难以调整

% 给出本文的方法，分析其中变量之间的内在关系，减小搜索空间。以及mapping的方式提前找到最佳kernel配置，在推理前先配置kernel再进行推理。

\subsection{Adaptive Kernel Search}

As we introduced in the previous subsection, the computational flow has many tunable hyperparameters, such as $w/x$ block width, warp row/col tiles, and so on. \rTwoMSB{The configuration of these hyperparameters is crucial for performance.}
% For example, as shown in Fig. \ref{fig:prefill kernel}, the APNN method with a fixed hyperparameter configuration exhibits up to a 50\% performance gap between W2A2 and W1A2. In contrast, APT benefits from flexible hyperparameter tuning, resulting in a performance gap of less than 16\%.}
% This also shows that there is no universal hyperparameter configuration that can achieve optimal performance for all types of MatMul operations. 

Take $w/x$ block width as an example: for square matrix GEMM, both hyperparameters should be as large as possible simultaneously, so that as much weight and activation data as possible can be stored in SHMEM. In contrast, for GEMV operations, due to the significant disparity in the size of the weight and feature matrices, and since the feature matrix is effectively a vector, setting \rOneMSB{$B_M$} to greater than 1 would cause memory redundancy in SHMEM. Thus, the value of $x$ block width should be significantly smaller than $w$ block width. Similarly, the optimal configuration for other hyperparameters varies significantly depending on the specific computational workload. When numerous hyperparameters need to be adjusted simultaneously, it is difficult to devise an empirical configuration strategy for manual tuning. 

Moreover, in LLM inference, there are significant differences between stages (such as GEMM in the prefill phase and GEMV in the decode phase), as well as substantial variations in matrix shapes across different locations (e.g., attention and MLP layers). Additionally, input differences and variations in the shape of feature matrices across different inference tasks further complicate the situation. These factors collectively make it impossible to rely on a fixed hyperparameter configuration to achieve optimal performance for APT. Therefore, we need an automatic operator optimization method that can dynamically find the best hyperparameter configuration to maximize inference performance.

Before kernel searching, to reduce the search space and ensure correct computation without memory overflow, we start by establishing the mathematical relationships between the tunable hyperparameters. As shown in Fig. \ref{fig:overview-APT}, each warp tile invokes \rOneMSB{$T_R\times T_C$} GEMM kernels. A single block consists of \rOneMSB{$W_B$} warps, and each GEMM kernel computes $W_M\times W_N$ outputs. Therefore, the relationships between the hyperparameters are as follows:
% \begin{equation}
%     block\_w\_bit\times block\_x\_bit=\newline
%     WM\times WN \times WPB \times WRT \times WCT
% \end{equation}
% \begin{equation}
% \begin{aligned}
    $$\frac{B_w\times B_x}{W_M\times W_N} = W_B \times T_R \times T_C.$$
% \end{aligned}
% \end{equation}
For each matrix operation requiring optimization, we traverse all hyperparameter combinations that satisfy the above mathematical relationships, based on the matrix size and bit width, to identify the configuration with the best performance. We record the optimal configurations for a wide range of common MNK combinations in LLM inference and store them in a lookup table for quick access during inference.

For configurations not recorded in the table, we provide two solutions:

\begin{enumerate}
    \item \textbf{Best Kernel Search Module}: This module identifies the optimal hyperparameter configurations for all matrix operations before inference and updates the configuration table. While this incurs additional preprocessing time, it is well-suited for scenarios involving repeated outputs from the same model or generating long sequences. 
    \item \textbf{Approximate Matching}: For quick inference without preprocessing, we use the closest matching matrix hyperparameters in the table. Though this approach may not achieve absolute optimal performance, it delivers near-optimal acceleration without additional inference overhead, making it ideal for smaller task workloads.
\end{enumerate}

% \begin{itemize}
%     \item 结合LLM模型结构, 指出存在不同大小的矩阵乘法运算, 并解释为何需要不同配置的kernel
%     \item 介绍上述kernel中的可变参数以及搜索空间, 体现人工选择kernel难以找到最优配置也难以及时根据输入更新
%     \item 提出在模型推理前根据输入的矩阵形状自适应搜索合适的kernel进行矩阵计算
% \end{itemize}}
\section{Experimental Results} \label{sec:results}
\subsection{Experimental Setup}
% In this section, we evaluate the performance of our arbitrary precision acceleration method for LLMs.
% \rOneSHK{Our experiments are conducted on the NVIDIA RTX 3090 GPU within an Ubuntu 18.04 system, using CUDA-11.8 driver and CUTLASS-2.11 repository\cite{cutlass}.}
\rThreeMSB{In this section, we evaluate the performance of our arbitrary precision acceleration method for LLMs. Our primary experiments and analysis are conducted on the NVIDIA RTX 3090 GPU within an Ubuntu 18.04 system, using CUDA-11.8 driver and CUTLASS-2.11 repository\cite{cutlass}. Additionally, we provide experimental results on RTX 4090\cite{ada} and H800\cite{hopper} GPUs, analyzing the similarities and differences in performance compared to the RTX 3090 results. The performance specifications of these GPUs are compared in Table \ref{tab:spec-gpu}.}

% \begin{table}[htbp]
%     \caption{\rThreeMSB{GPU Specification Comparison}}
%     \vspace{+2.0 em}
%     \begin{tabular}{cccccc}
%     \hline
%     \textbf{Specification} & \textbf{Architecture} & \textbf{Bandwidth} & \textbf{INT MatMul} \\ \hline
%     \textbf{RTX 3090} & Ampere       & 936.2 GB/s         & INT8, INT4, INT1 \\ \hline
%     \textbf{RTX 4090} & Ada Lovelace & 1 TB/s & INT8, INT4, INT1 \\ \hline
%     \textbf{H800}     & Hopper       & 2   TB/s       & INT8, INT1 \\ \hline    
%     \end{tabular}
%     \label{tab:spec-gpu}
% \end{table}

\begin{table}[tbp]
    \caption{\rTwoFC{GPU Specification Comparison}}
    \vspace{+2.0 em}
    \centering
    \resizebox{\linewidth}{!}{%
        \begin{tabular}{cccccc}
        \hline
        \multirow{2}{*}{\textbf{Spec.}} & \multirow{2}{*}{\textbf{Architecture}} & \multirow{2}{*}{\textbf{\#TCs}} & \multirow{2}{*}{\textbf{Capacity}} & \textbf{Memory} & \textbf{Supported} \\
         &  &  &  & \textbf{Bandwidth} & \textbf{INT MatMul} \\
        \hline
        \textbf{RTX 3090} & Ampere & 328 & 24 GB & 936.2 GB/s & INT8, INT4, INT1 \\
        \hline
        \textbf{RTX 4090} & Ada Lovelace & 512 & 24 GB & 1 TB/s & INT8, INT4, INT1 \\
        \hline
        \textbf{H800} & Hopper & 528 & 80 GB & 2 TB/s & INT8, INT1 \\
        \hline
    \end{tabular}%
    }
    \vspace{-1em}
    \label{tab:spec-gpu}
\end{table}

\rOneMSB{We evaluate our designs, including W1A2 (1-bit weights and 2-bit activations), W2A2, and W3A4, against standard FP32 and FP16 MatMuls. On the RTX 3090, we benchmark them against NVIDIA’s \rTwoFC{TC}-accelerated ultra-low-bit MatMul implementations (CUTLASS INT1 and CUTLASS INT4) to highlight our superior computational performance on the same TCs. Finally, we compare our approach with other \rOneFC{state-of-the-art (SOTA)} TC-based MatMul acceleration methods, such as APNN-TC\cite{apnn}, BSTC\cite{bstc}, and BTC\cite{btc}, demonstrating the significant performance advantages of our work in low-bit arbitrary precision computation.}
% \rThreeMSB{On the RTX 4090 and H800, we limit our comparison to CUTLASS INT4/INT8 (as the Hopper architecture lacks direct support for INT4 MatMul\cite{hopper}) and APNN-TC\cite{apnn}.}
\rTwoFC{On the RTX 4090 and H800, we evaluate against APNN-TC on both platforms, along with CUTLASS INT4 on the RTX 4090 and CUTLASS INT8 on the H800, since the Hopper architecture does not natively support INT4 operations.}

The evaluation is divided into \rOneMSB{three} main parts: (1) an assessment of our arbitrary precision MatMul kernels, (2) an analysis of its impact on LLM inference performance, and \rOneMSB{(3) an evaluation for the acceleration effect of APT's kernel mapping strategy}. 
% Through these experiments, w
We aim to demonstrate the effectiveness of our approach in accelerating LLM computations across \rOneSHK{basic operator and entire model levels.}
% \rOneFC{All results reported are the average of \TODO{XXX} times execution.}

\subsection{Arbitrary Precision Kernel Evaluation}
In this section, we evaluate the computational performance of our arbitrary-precision MatMul kernel design across various MatMul tasks. 
First, we assess the speedup of our design in MatMul tasks compared to other implementation methods. Then, to evaluate its performance in LLMs, we extract key MatMul operations from the LLAMA3-8B, and evaluate the prefill and decode phases separately.
% Finally, we design an ablation study to demonstrate the effectiveness of our memory management and kernel mapping strategies.

\subsubsection{\textbf{General MatMul}}
To evaluate general MatMul, we assume that the weight matrix is a square matrix (i.e. $N=K$), while the input matrix has a fixed dimension of $M=64$. \rOneMSB{This simplification allows for more straightforward performance comparison across different approaches while still effectively demonstrating the computational efficiency of our method for typical LLM operations.} Under this condition, we compare the MatMul performance of APT with other baseline methods.

\begin{table}[tb]\fontsize{5.8pt}{7.2pt}\selectfont
        \vspace{-1.5em}
	\centering
        \caption{General MatMuls on RTX 3090}
        \vspace{2.5em}
	\begin{tabular}{|c|c|c|c|c|c|c|}
		\hline
		\textbf{M/N/K}        & \multicolumn{2}{c|}{\textbf{64/1k/1k}}     & \multicolumn{2}{c|}{\textbf{64/2k/2k}}     & \multicolumn{2}{c|}{\textbf{64/4k/4k}}     \\ \hline
		\textbf{Schemes}      & \textbf{Latency}         & \textbf{Speedup}         & \textbf{Latency}         & \textbf{Speedup}         & \textbf{Latency}         & \textbf{Speedup}         \\ \hline
		\textbf{FP32}         & 33.7us         & 1.00$\times$          & 49.3us         & 1.00$\times$          & 119us          & 1.00$\times$          \\ \hline
		\textbf{FP16}         & 19.4us         & 1.74$\times$          & 31.6us         & 1.56$\times$          & 53.7us          & 2.21$\times$          \\ \hline
            \textbf{CUTLASS INT1} & 9.19us         & 3.67$\times$          & 11.8us         & 4.18$\times$          & 17.2us         & 6.92$\times$          \\ \hline
		\textbf{CUTLASS INT4} & 15.6us         & 2.16$\times$          & 25.3us         & 1.95$\times$          & 47.4us         & 2.51$\times$          \\ \hline \hline
		\textbf{APT W3A4}     & 11.6us         & 2.91$\times$          & 21.5us         & 2.29$\times$          & 42.3us         & 2.81$\times$          \\ \hline 
		\textbf{APT W2A2}     & 8.40us         & 4.01$\times$          & 11.8us         & 4.18$\times$          & 20.7us         & 5.75$\times$          \\ \hline
		\textbf{APT W1A2}     & 6.83us         & 4.93$\times$          & 11.6us         & 4.25$\times$          & 18.6us         & 6.40$\times$          \\ \hline
	\end{tabular}
	\label{tab:gemm}
        \vspace{-1em}
\end{table}

\begin{figure}[tb]
	\centering
	\includegraphics[width=\linewidth]{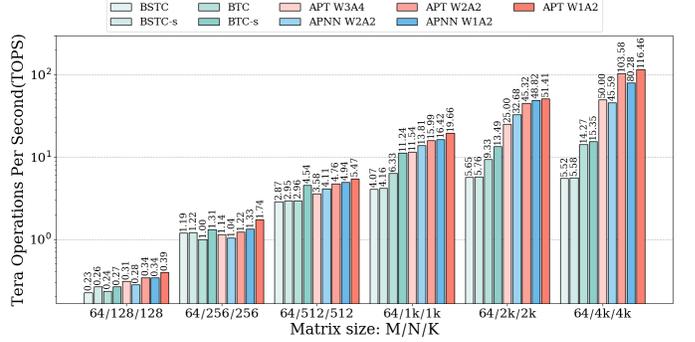}
	\caption{Comparison of throughput between APT and other methods in the context of general MatMuls on RTX 3090.}
	\label{fig:gemm}
        \vspace{-1.5em}
\end{figure}

{\textbf{Comparisons with FP and CUTLASS baselines:}} Table \ref{tab:gemm} presents a performance comparison between our APT, PyTorch floating-point MatMul, and NVIDIA CUTLASS low-bit INT MatMul for the MatMul task. The latency is measured as the average processing time for 1000 computations, and the speedup is referenced against FP32 MatMul.

Compared to PyTorch FP32 and FP16, APT achieves significant acceleration for MatMul. Under the W1A2 and W2A2 configurations, APT achieves over 4$\times$ speedup compared to FP32 and more than 2.3$\times$ speedup compared to FP16. Specifically, under the W1A2 configuration, APT achieves up to 6.40$\times$ speedup over FP32 and 2.90$\times$ speedup over FP16.

For the CUTLASS kernel, due to limited support for data formats, only INT4 and INT1 low-precision MatMul kernels are available. It can be observed that, with the advantages of TC acceleration and low-precision, the CUTLASS kernel achieves substantial speedup compared to floating-point operations. Compared to CUTLASS INT4, APT significantly outperforms in all three configurations. This is largely due to its support for arbitrary-precision MatMul, as without APT, these configurations could only use CUTLASS INT4 for computation. When compared to CUTLASS INT1, even though the computational bit-width is not advantageous, in the 64/1k/1k and 64/2k/2k tasks, the W1A2 and W2A2 configurations both outperform CUTLASS INT1. In the 64/4k/4k task, the performance of APT W1A2 approaches that of CUTLASS INT1, achieving a 92.5\% speedup. This indicates that the advantages of APT stem not only from the precision savings enabled by arbitrary precision but also from its efficient memory management and optimal operator search strategy.

% {\textbf{Comparisons with related works:}} 
{\textbf{Comparisons with \rOneFC{prior SOTAs}:}} Fig. \ref{fig:gemm} presents a performance comparison of APT against other similar works across MatMul tasks of varying sizes using Tera Operations Per Second (TOPS) as the metric. Specifically, BSTC refers to the 32-bit BSTC BMM in \cite{bstc}, while BSTC-s denotes the fine-grained 32-bit BSTC BMM in \cite{bstc}. BTC represents the basic BTC implementation in \cite{btc}, and BTC-s refers to BTC with 128-bit load and shared memory optimization in \cite{btc}. Note that APNN-TC is compared only with W1A2 and W2A2 configurations due to its limited precision support.

% Since both BSTC and BTC were optimized for earlier NVIDIA GPU architectures, their performance on the Ampere architecture is suboptimal. Specifically, BSTC does not utilize \rTwoFC{TC}s for matrix computation, resulting in \rOneMSB{(more than 20$\times$ slower than APT W1A2 at most)}.
\rTwoFC{Both BSTC and BTC, being optimized for earlier NVIDIA GPU architectures, perform suboptimally on Ampere. In particular, BSTC's lack of TC utilization
% for matrix computation 
results in significantly degraded performance, being more than 20$\times$ slower than APT W1A2 in some cases.}
While BTC leverages \rTwoFC{TC}s, it lacks proper adaptation to the Ampere architecture. 
\rOneMSB{Compared to APNN-TC, our proposed APT offers enhanced flexibility for arbitrary-precision matrix computations (e.g., APT supports W3A4 configurations that APNN-TC cannot process) while also implementing more efficient memory scheduling through dual shared memory allocation for separate data loading and computation processes.}
Additionally, APT employs an optimized kernel mapping strategy to further enhance performance. \rOneMSB{As a result, for the same W2A2 configuration, APT achieves up to 2.27$\times$ faster performance than APNN-TC.}

\begin{table}[t]\fontsize{5.8pt}{7.2pt}\selectfont
        \vspace{-1.5em}
	\centering
        \caption{LLM Prefill MatMul on RTX 3090 (GEMM)}
        \vspace{2.5em}
	\begin{tabular}{|c|c|c|c|c|c|c|}
		\hline
		\textbf{M/N/K}        & \multicolumn{2}{c|}{\textbf{64/1k/4k}}     & \multicolumn{2}{c|}{\textbf{64/14k/4k}}     & \multicolumn{2}{c|}{\textbf{64/4k/14k}}     \\ \hline
		\textbf{Schemes}      & \textbf{Latency}         & \textbf{Speedup}         & \textbf{Latency}         & \textbf{Speedup}         & \textbf{Latency}         & \textbf{Speedup}         \\ \hline
		\textbf{FP32}         & 41.6us         & 1.00$\times$          & 654us         & 1.00$\times$          & 434us          & 1.00$\times$          \\ \hline
		\textbf{FP16}         & 39.8us         & 1.05$\times$          & 302us         & 2.17$\times$          & 164us          & 2.65$\times$          \\ \hline
		\textbf{CUTLASS INT4} & 88.6us         & 0.47$\times$          & 143us         & 4.57$\times$          & 291us          & 1.49$\times$          \\ \hline
		\textbf{CUTLASS INT1} & 29.8us         & 1.40$\times$          & 44.4us        & 14.7$\times$          & 88.6us         & 4.90$\times$          \\ \hline \hline
		\textbf{APT W3A4}     & 26.8us         & 1.55$\times$          & 203us         & 3.22$\times$          & 252us          & 1.72$\times$          \\ \hline 
		\textbf{APT W2A2}     & 15.7us         & 2.64$\times$          & 62.4us        & 10.5$\times$          & 71.7us         & 6.05$\times$          \\ \hline
		\textbf{APT W1A2}     & 15.9us         & 2.62$\times$          & 52.8us        & 12.4$\times$          & 63.6us         & 6.82$\times$          \\ \hline
	\end{tabular}
        \vspace{-1em}
	\label{tab:llm-prefill}
\end{table}

\begin{figure}[tb]
	\centering
	\includegraphics[width=\linewidth]{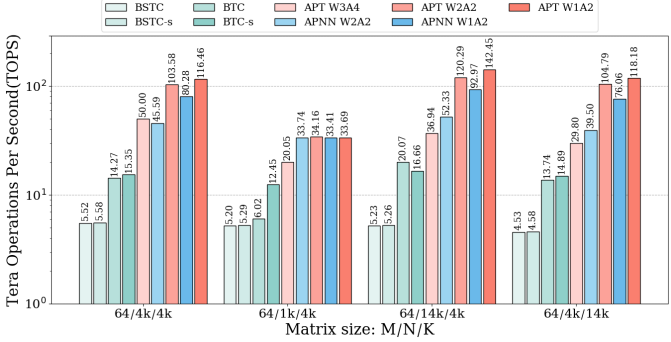}
	\caption{Comparison of throughput between APT and other methods in LLM prefill phase MatMul tasks on RTX 3090.}
        \label{fig:prefill kernel}
        \vspace{-1.5em}
\end{figure}

\subsubsection{\textbf{MatMul in LLMs}}\label{sec:llm-matmul}
In addition to testing the general MatMul performance, to more accurately reflect the performance of APT MatMul in LLM inference, we extract linear layer configurations with the highest latency from the LLAMA3-8B model\rOneMSB{, as shown in Fig. \ref{fig:matmul different},} and measure the performance comparison between APT and the baselines for these tasks. We separately evaluate the MatMul performance during the prefill and decode stages.

\textbf{MatMuls in prefill phase:}  We assume an input token length of 64, i.e., $M=64$. Table \ref{tab:llm-prefill} presents a comparison of APT with PyTorch floating-point MatMul and CUTLASS low-precision kernels. In the case of PyTorch's floating-point MatMul, the two large, compute-intensive MatMuls result in \rOneMSB{0.16-0.65ms} latency.
% , severely impacting the model inference speed. 
In contrast, APT's W2A2 and W1A2 configurations achieve more than 10$\times$ and 6$\times$ speedup compared to FP32 MatMul for these two matrices. 
% This demonstrates that TC exhibits a significant advantage for more compute-intensive tasks. 
For CUTLASS, since it also utilizes TC for acceleration, the speedup for the two large MatMuls is similarly significant. However, the MatMul speed of all three APT configurations significantly surpasses that of the CUTLASS INT4 kernel, with the W1A2 configuration achieving a speedup of 2.7-5.5$\times$. As for the CUTLASS INT1 kernel, even when the bit-width does not have an advantage, both the W1A2 and W2A2 configurations approach or even exceed its inference speed. Specifically, the total latency for the three matrix calculations in W1A2 is 1.23$\times$ faster than CUTLASS INT1.

% {\textbf{Comparisons with related works:}} 
{\textbf{Comparisons with \rOneFC{prior SOTAs}:}} Fig. \ref{fig:prefill kernel} presents a performance comparison between APT and other SOTAs across four primary MatMul tasks during the prefill phase of the LLAMA3-8B model. Compared to APNN-TC, APT does not exhibit a significant performance improvement in smaller tasks (e.g., 64/1k/4k). However, for larger MatMul tasks, APT achieves over 1.5$\times$ speedup under W1A2 compared to APNN-TC with the same configuration, and up to 2.65$\times$ acceleration under W2A2. This substantial speedup is due to APT’s adaptive kernel configurations tailored to different computation tasks, whereas APNN-TC lacks the flexibility to adjust its kernel settings dynamically. 
% As a result, APT achieves remarkable acceleration in certain tasks by selecting more optimal kernel configurations.

\begin{table}[t]\fontsize{5.8pt}{7.2pt}\selectfont
        \vspace{-1.5em}
	\centering
        \caption{LLM Decode MatMul on RTX 3090 (GEMV)}
        \vspace{2.5em}
	\begin{tabular}{|c|c|c|c|c|c|c|}
		\hline
		\textbf{M/N/K}        & \multicolumn{2}{c|}{\textbf{1/1k/4k}}     & \multicolumn{2}{c|}{\textbf{1/14k/4k}}     & \multicolumn{2}{c|}{\textbf{1/4k/14k}}     \\ \hline
		\textbf{Schemes}      & \textbf{Latency}         & \textbf{Speedup}         & \textbf{Latency}         & \textbf{Speedup}         & \textbf{Latency}         & \textbf{Speedup}         \\ \hline
		\textbf{FP32}         & 48.0us         & 1.00$\times$          & 330us         & 1.00$\times$          & 339us          & 1.00$\times$          \\ \hline
		\textbf{FP16}         & 35.9us         & 1.34$\times$          & 298us         & 1.11$\times$          & 162us          & 2.09$\times$          \\ \hline
		\textbf{CUTLASS INT4} & 44.8us         & 1.07$\times$          & 67.4us         & 4.90$\times$          & 265us         & 1.28$\times$          \\ \hline
		\textbf{CUTLASS INT1} & 16.8us         & 2.86$\times$          & 38.4us         & 8.59$\times$          & 44.5us         & 7.62$\times$          \\ \hline \hline
		\textbf{APT W3A4}  & 21.4us         & 2.24$\times$          & 161us         & 2.05$\times$          & 216us         & 1.57$\times$          \\ \hline 
		\textbf{APT W2A2}  & 15.1us         & 3.18$\times$          & 51.2us         & 6.45$\times$          & 60.1us         & 5.64$\times$          \\ \hline
		\textbf{APT W1A2}  & 15.9us         & 3.02$\times$          & 53.0us         & 6.23$\times$          & 64.1us         & 5.29$\times$          \\ \hline
	\end{tabular}
        \vspace{-1em}
	\label{tab:llm-decode}
\end{table}

\begin{figure}[tb]
	\centering
	\includegraphics[width=\linewidth]{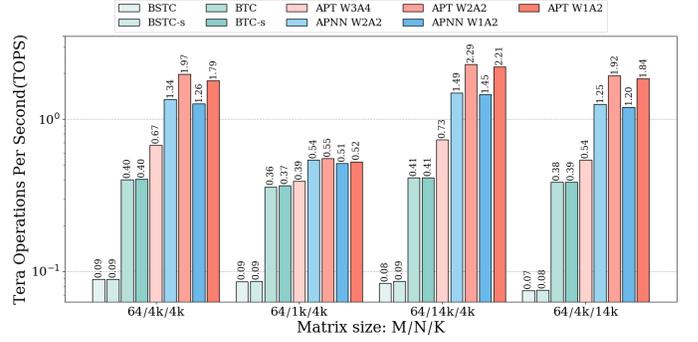}
	\caption{Comparison of throughput between APT and other methods in LLM decode phase MatMul tasks on RTX 3090.}
	\label{fig:decode kernel}
        \vspace{-1.5em}
\end{figure}

\textbf{MatMuls in decode phase:} In this case, the input matrix is transformed into a vector (i.e., $M=1$). Table~\ref{tab:llm-decode} presents the performance of APT, PyTorch floating-point, and CUTLASS kernels in MatMul during the decode stage of the LLAMA3-7B model. For CUTLASS, as mentioned earlier, during TC computation, a 8/8/128 MatMul is processed at once. However, for matrix-vector multiplication, since the input matrix is a vector, it must be padded with seven zero vectors to invoke TC, leading to a performance loss of 87.5\% of TC’s potential. 
% As a result, CUTLASS INT1’s speedup is far less significant than in the MatMul task, and CUTLASS INT4 is even slower than FP32. 

However, bit-level decomposition, APT is able to better utilize the performance of TC. It is worth noting that, while APT partially addresses this issue, W1A2 still loses some performance, resulting in slower computation speeds compared to W2A2. Compared to FP32, APT W2A2 achieves up to 6.45$\times$ speedup, with the total latency of the three matrix MatMuls reduced by 5.67$\times$. Compared to FP16, W2A2 achieves up to 5.81$\times$ speedup in the 1/14k/4k matrix MatMul, with an overall speedup of 3.92$\times$. Compared to CUTLASS, both W1A2 and W2A2 significantly outperform CUTLASS INT4, and even with a bit-width disadvantage, they approach the speed of CUTLASS INT1.

% {\textbf{Comparisons with related works:}} 
{\textbf{Comparisons with \rOneFC{prior SOTAs}:}} Fig. \ref{fig:decode kernel} shows a TOPS comparison between APT and other prior works in the primary MatMul tasks during the decode phase of the LLAMA3-8B model. It can be observed that in GEMV tasks, the performance gap between BTC and APT is significantly reduced. This is because BTC also utilizes 1-bit \rTwoFC{TC}s as its primary computation resource. However, since BTC was optimized for the Turing architecture, performing large-scale matrix computations on Ampere architecture evidently encounters a roofline bottleneck. Compared to APNN-TC, APT achieves over 1.5$\times$ speedup in ultra-large MatMul tasks under both W1A2 and W2A2 configurations.

\subsubsection{\textbf{MatMuls on Different GPU Platforms}}

\begin{figure*}[tb]
    \centering
    \subfigure[General MatMul on RTX 4090]{
        \includegraphics[width=0.31\linewidth]{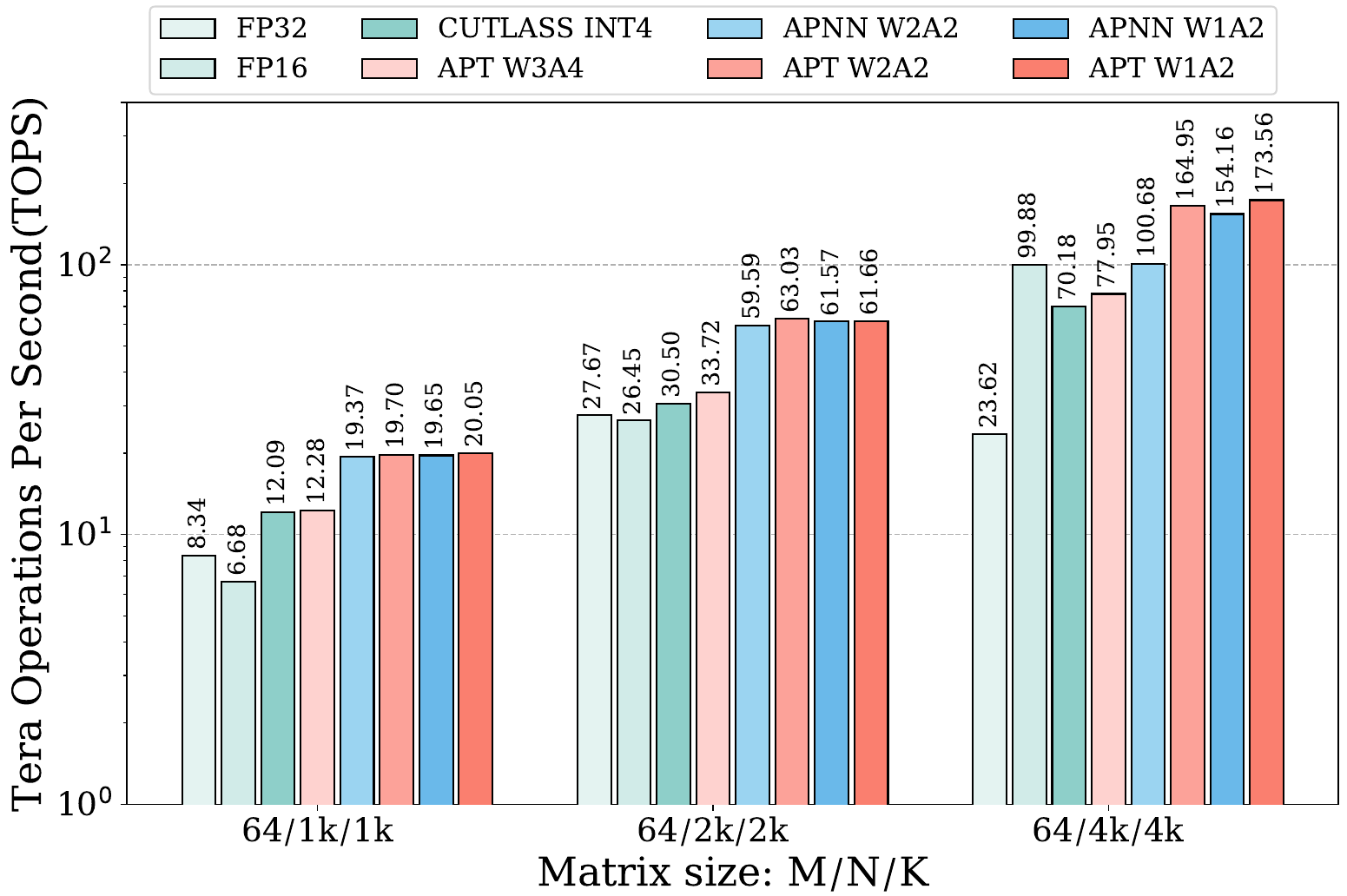}
        \label{subfig:4090-gemm}
    }
    \subfigure[Prefill MatMul tasks on RTX 4090]{
        \includegraphics[width=0.31\linewidth]{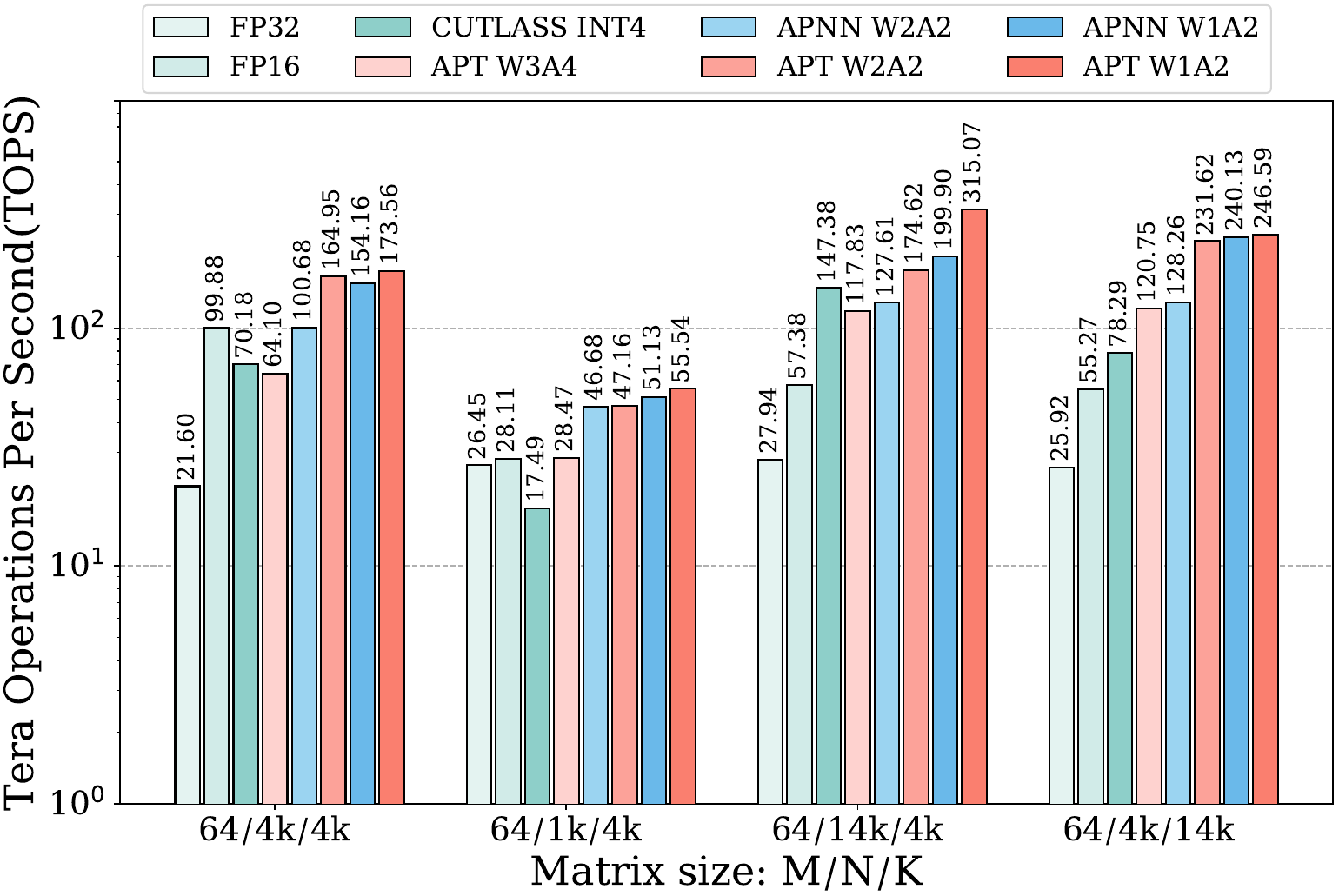}
        \label{subfig:4090-llm-gemm}
    }
    \subfigure[Decode MatMul tasks on RTX 4090]{
        \includegraphics[width=0.31\linewidth]{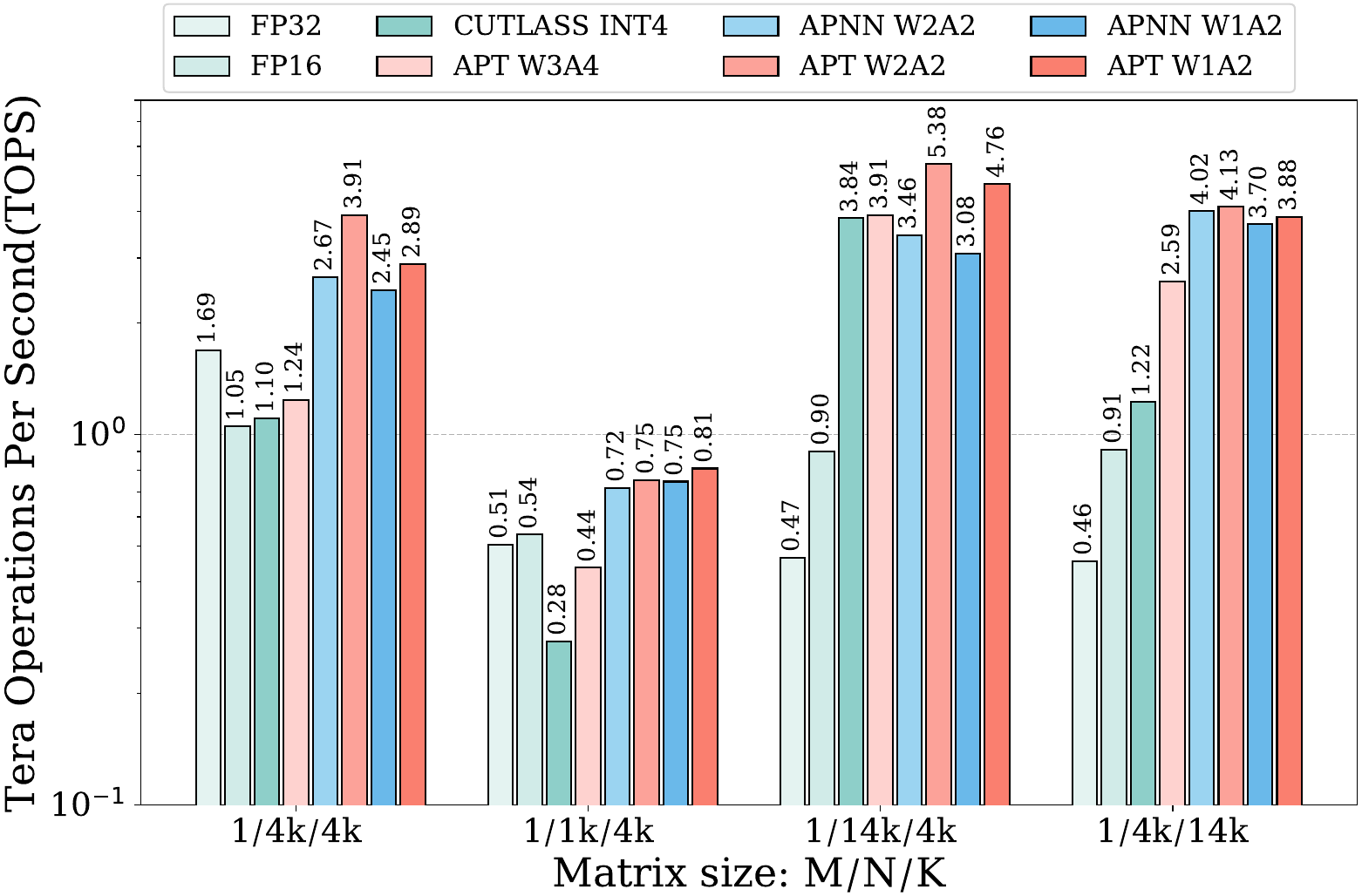}
        \label{subfig:4090-llm-gemv}
    }
    \subfigure[General MatMul on H800]{
        \includegraphics[width=0.31\linewidth]{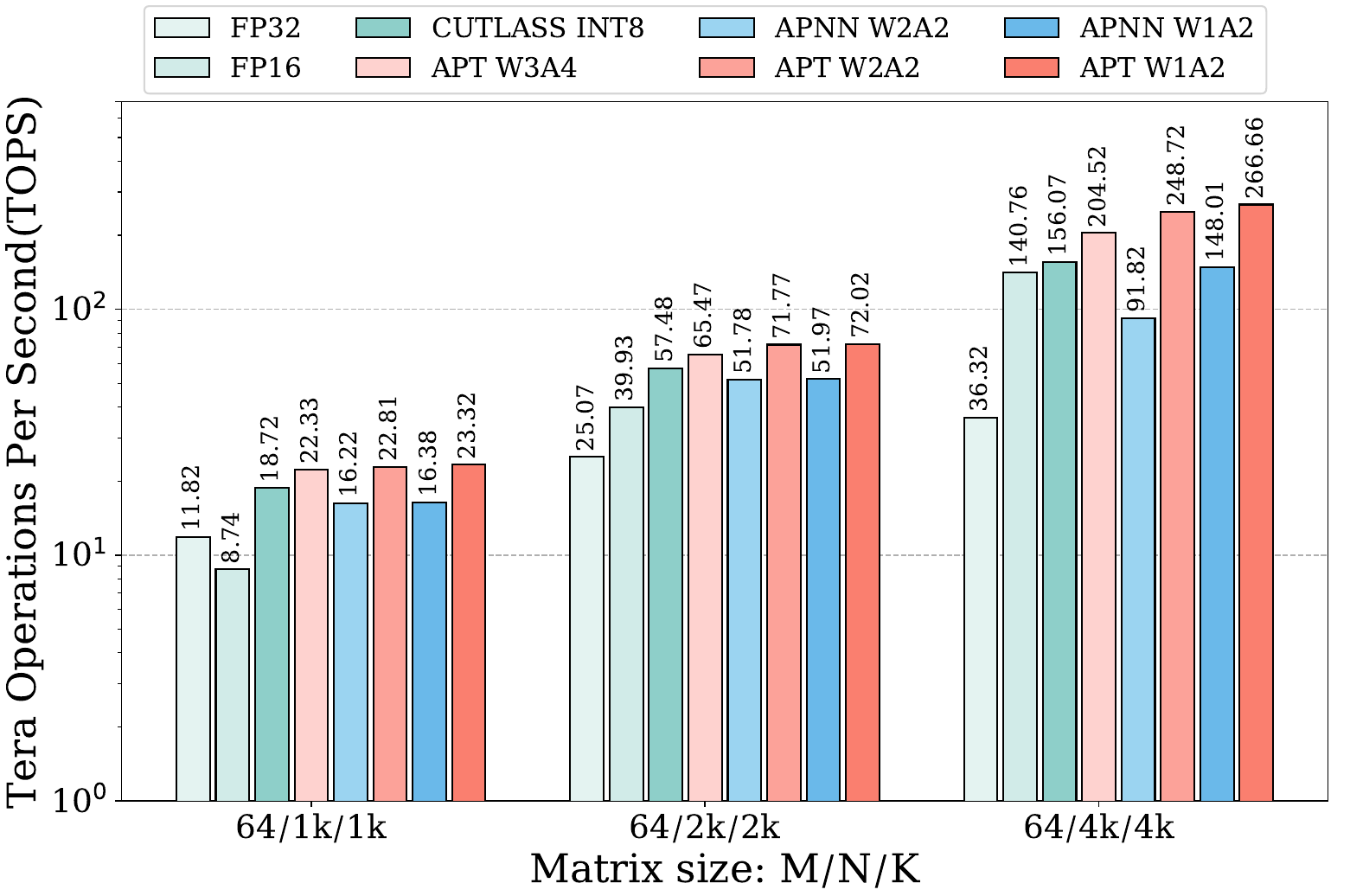}
        \label{subfig:h800-gemm}
    }
    \subfigure[Prefill MatMul tasks on H800]{
        \includegraphics[width=0.31\linewidth]{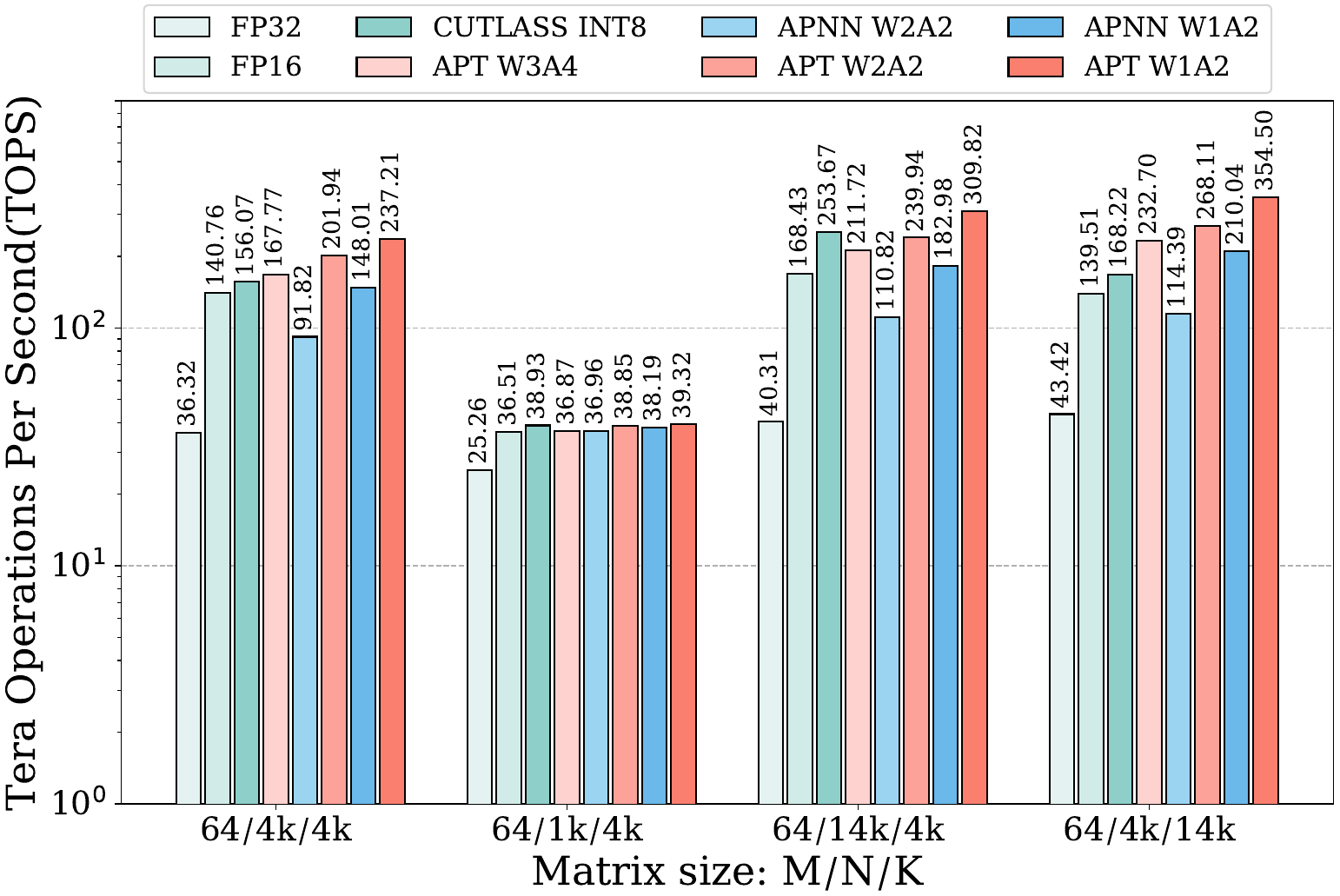}
        \label{subfig:h800-llm-gemm}
    }
    \subfigure[Decode MatMul tasks on H800]{
        \includegraphics[width=0.31\linewidth]{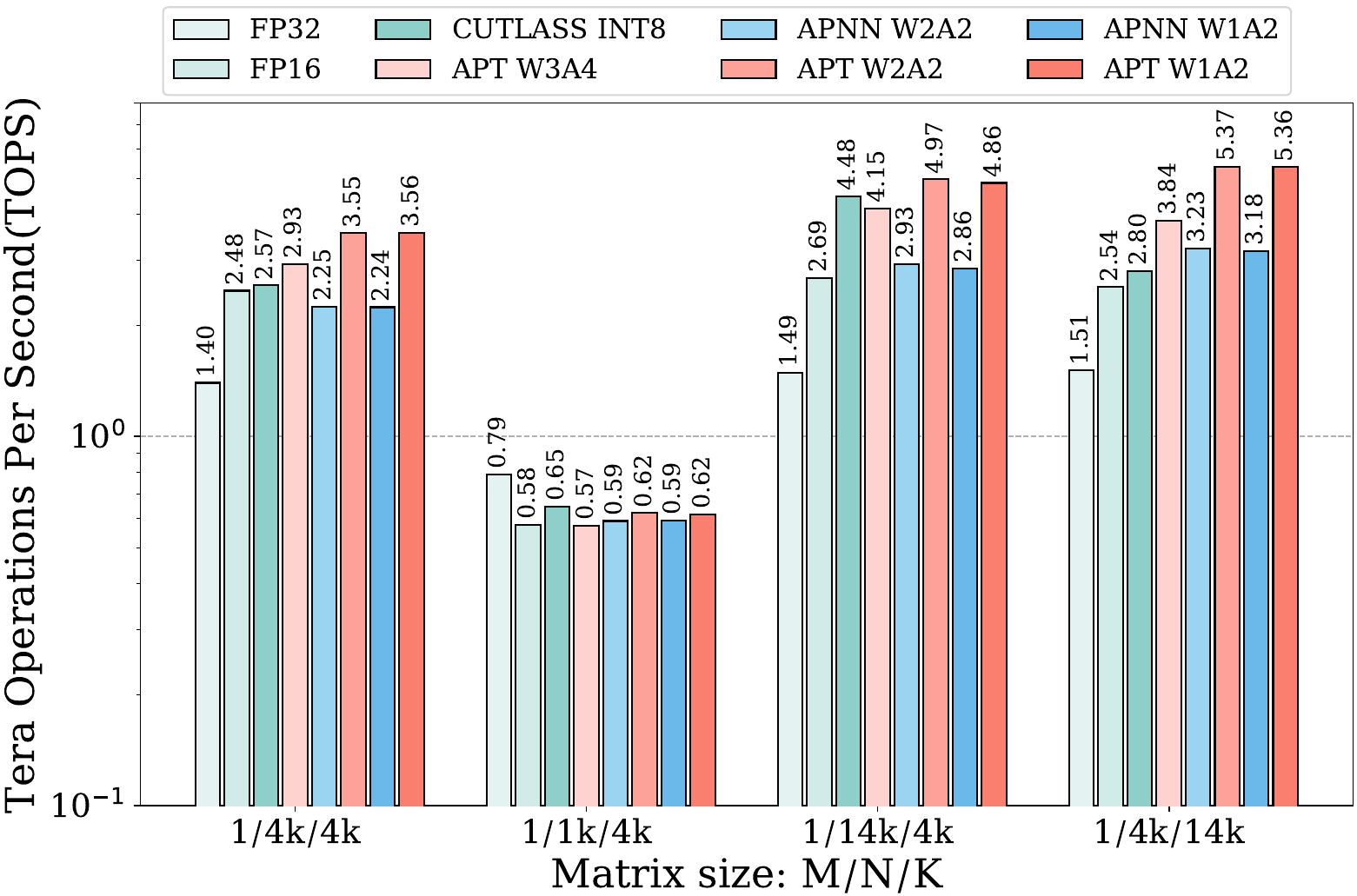}
        \label{subfig:h800-llm-gemv}
    }
    \vspace{+2em}
    \caption{\rThreeMSB{Comparison of throughput between APT and other methods on different GPU platforms.}}
    \label{fig:matmul-gpus}
    \vspace{-1em}
\end{figure*}

\rThreeMSB{In addition to the experiments conducted on the RTX 3090, we extend our performance evaluation to include the RTX 4090 and H800 GPUs. The detailed results are presented in Fig. \ref{fig:matmul-gpus}. For RTX 4090 (Ada Lovelace architecture), our APT-LLM achieves up to 5.98× speedup over FP16 and 3.37× over CUTLASS INT4, as shown in Fig. \ref{subfig:4090-llm-gemv}. For H800 (Hopper architecture), our APT-LLM achieves up to 2.54× speedup over FP16 and 2.10× over CUTLASS INT8, as shown in Fig. \ref{subfig:h800-llm-gemm}. In brief, our method consistently demonstrates superior performance across all three platforms when compared to FP32 and FP16 MatMuls, as well as optimized low-bit implementations such as CUTLASS INT4/INT8 and APNN-TC.}

\rThreeMSB{However, the observed speedup ratios on the newer RTX 4090 and H800 GPUs, when compared to the FP baselines and CUTLASS implementations, are slightly less pronounced than those observed on the RTX 3090. This difference can be attributed to several factors: (1) Ada Lovelace (RTX 4090) and Hopper (H800) architectures have improved low-bit INT \rTwoFC{TC} performance compared to Ampere, making CUTLASS INT4 and INT8 more competitive; (2) the enhanced memory bandwidth and cache hierarchy in Ada Lovelace and Hopper may reduce the relative advantage of our shared-memory-centric optimization strategy; and (3) different architectural optimizations across GPU generations may affect the effectiveness of our bit-fragmentation approach differently. Nonetheless, the absolute performance gains measured in TOPS clearly indicate that our method continues to offer significant enhancements in computational throughput on the latest GPU architectures.}

\subsection{Arbitrary Precision LLM Evaluation}

% \begin{figure*}[tbp]
% 	\centering
% 	\includegraphics[width=\linewidth]{figures/model-prefill.pdf}
%         \vspace{+0.5em}
% 	\caption{Comparison of APT and other methods in accelerating LLM inference during the prefill stage.}
% 	\label{fig:model-prefill}
% \end{figure*}
% \begin{figure*}[tbp]
% 	\centering
% 	\includegraphics[width=\linewidth]{figures/model-decode.pdf}
%         \vspace{+0.5em}
% 	\caption{Comparison of APT and other methods in accelerating LLM inference during the decode stage.}
%         \vspace{-1em}
% 	\label{fig:model-decode}
% \end{figure*}
\begin{figure*}
    \centering
    \subfigure[Prefill stage]{
        \includegraphics[width=\linewidth]{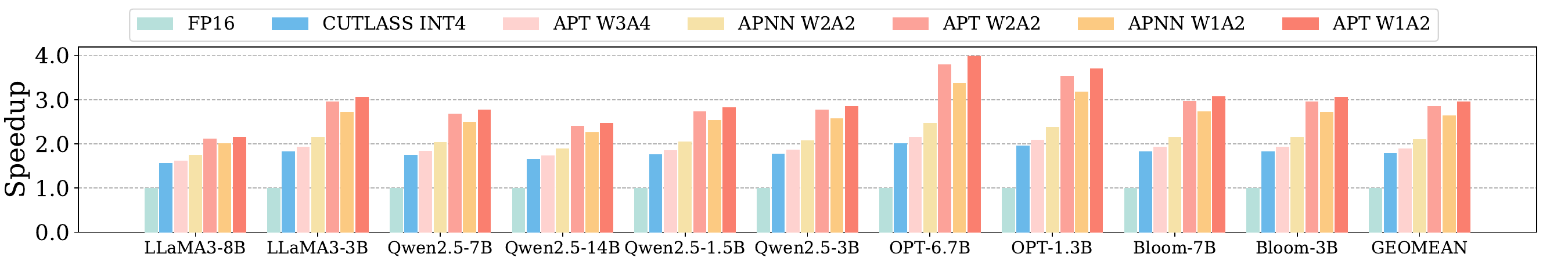}
	\label{fig:model-prefill}
    }
    % \vspace{-2em}
    \subfigure[Decode stage]{
        \includegraphics[width=\linewidth]{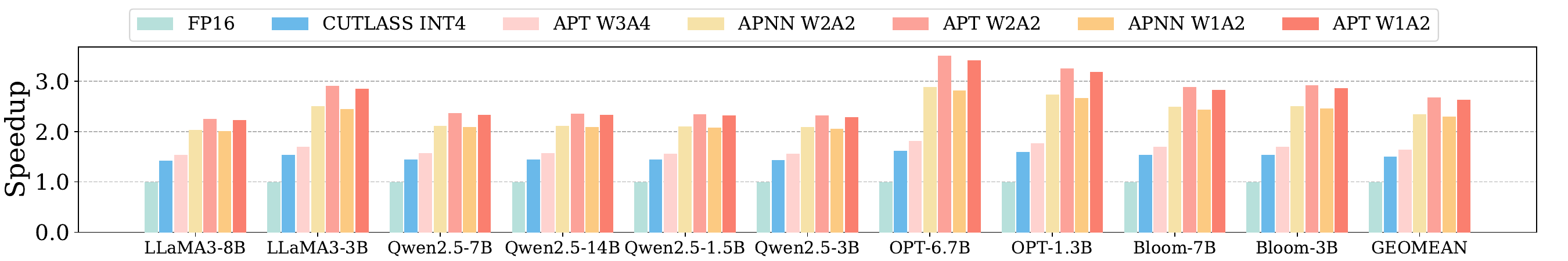}
	\label{fig:model-decode}
    }
    \vspace{+1em}
    \caption{\rThreeMSB{Comparison of APT and other methods in accelerating LLM inference on RTX 3090.}}
    \label{fig:model-eval}
    \vspace{-1em}
\end{figure*}

% Efficient inference of LLMs is crucial for their practical deployment. 
% We evaluate our arbitrary precision MatMul design in ultra-low bit quantized LLMs to demonstrate its effectiveness in accelerating real-world LLM inference.

In this section, we evaluate APT's inference acceleration across various LLMs under different precision settings. The tested models include: LLaMA3\cite{llama-3}, Qwen2.5\cite{yang2024qwen2}, OPT\cite{zhang2022opt} and Bloom\cite{workshop2022bloom}. We use the GPTQ quantization method\cite{gptq} to quantize these models into low-bit representations and replace their MatMul operations with our APT kernel and APNN-TC kernel. The inference speed of these models is then compared against their FP16-based counterparts to measure the speedup achieved by APT-LLM. Additionally, we quantize these models to 4-bit precision using GPTQ and perform matrix computations with CUTLASS INT4, which serves as the GPU-accelerated baseline for comparison.

% LLM inference consists of two phases: prefill and decode. 
We evaluate APT-LLM's acceleration performance for both phases on a single NVIDIA RTX 3090 GPU. Specifically, we measure the speedup per token generation for prefill phase ($M=64$) and decode phase ($M=1$).

\subsubsection{\textbf{Prefill phase}}
Fig. \ref{fig:model-prefill} presents a comparison of APT-LLM's acceleration performance across different precision settings during the prefill phase for various LLMs. 
% The results demonstrate the speedup achieved by APT-LLM relative to the baseline FP16 inference method and the CUTLASS INT4 acceleration approach, which utilizes W4A4 quantization. 
% By evaluating models of different sizes and architectures, the figure highlights how APT-LLM effectively optimizes MatMul operations to achieve superior inference efficiency under various precision configurations. 
The results demonstrate that APT-LLM's inference performance across all three precision settings consistently outperforms both PyTorch FP16 and CUTLASS INT4. Among the tested models, the OPT-6.7B model exhibits the most significant improvement, achieving a 3.99$\times$ speedup over FP16 and a 1.98$\times$ speedup over CUTLASS. On average, APT-LLM delivers a 2.96$\times$ acceleration compared to FP16 and a 1.65$\times$ acceleration compared to CUTLASS, highlighting its efficiency in optimizing MatMul operations for large-scale model inference.

\subsubsection{\textbf{Decode phase}}

Fig. \ref{fig:model-decode} presents a comparison of APT-LLM's acceleration performance across different precision settings during the decode phase of inference for various LLMs. 
% The results demonstrate the speedup achieved by APT-LLM relative to the baseline FP16 inference method and the CUTLASS INT4 acceleration approach, which utilizes W4A4 quantization. 
% By evaluating models of different sizes and architectures, the figure highlights how APT-LLM effectively optimizes MatMul operations to achieve superior inference efficiency under various precision configurations.
As mentioned in the Sec. \ref{sec:llm-matmul}, the utilization of \rTwoFC{TC}s is relatively low during matrix-vector multiplication, which results in less significant acceleration in the decode phase compared to the prefill phase. However, APT-LLM still achieves up to 3.50$\times$ speedup over FP16, with an average speedup of 2.68$\times$. Compared to CUTLASS INT4, APT-LLM leverages \rTwoFC{TC} resources more effectively, partially mitigating this issue. As a result, the acceleration in the decode phase is even more pronounced, reaching a maximum speedup of 2.16$\times$ and an average speedup of 1.78$\times$.

\subsubsection{\textbf{LLM Evaluation on Different GPU Platforms}}

\begin{figure*}[tbp]
    \centering
    \subfigure[Prefill on RTX 4090]{
        \includegraphics[width=0.48\linewidth]{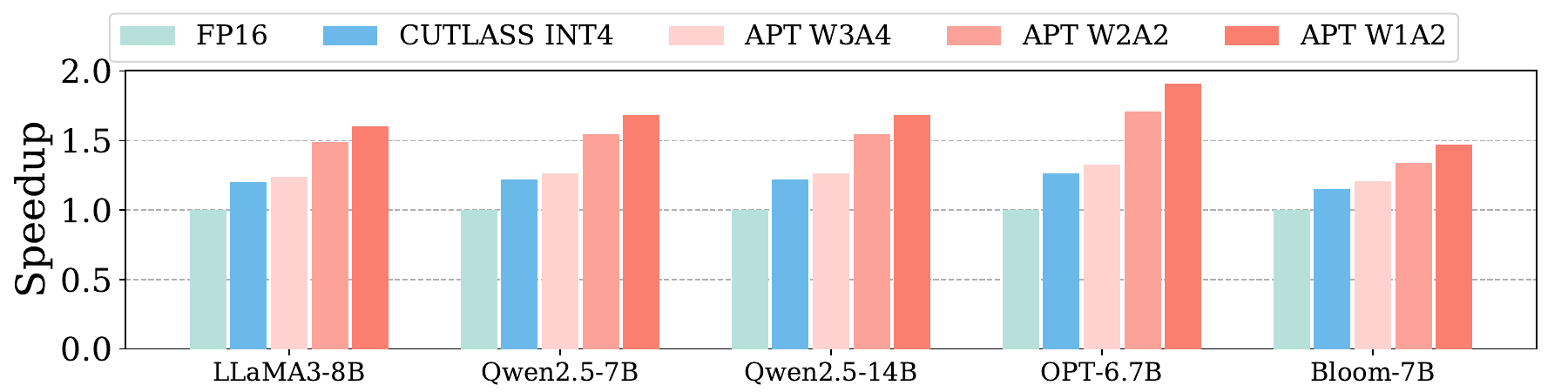}
        \label{fig:model-prefill-4090}
    }
    \subfigure[Decode on RTX 4090]{
        \includegraphics[width=0.48\linewidth]{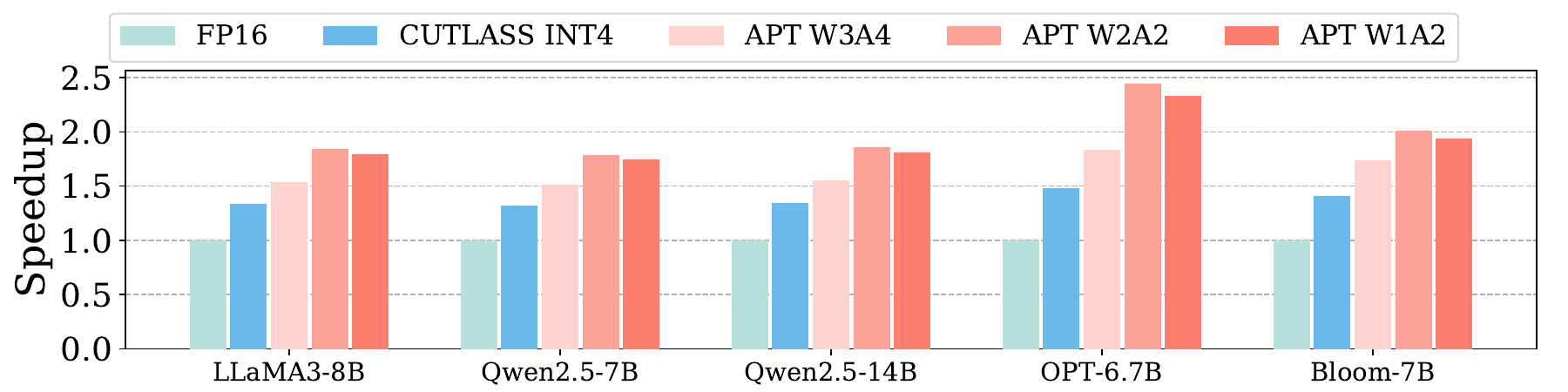}
        \label{fig:model-decode-4090}
    }
    \subfigure[Prefill on H800]{
        \includegraphics[width=0.48\linewidth]{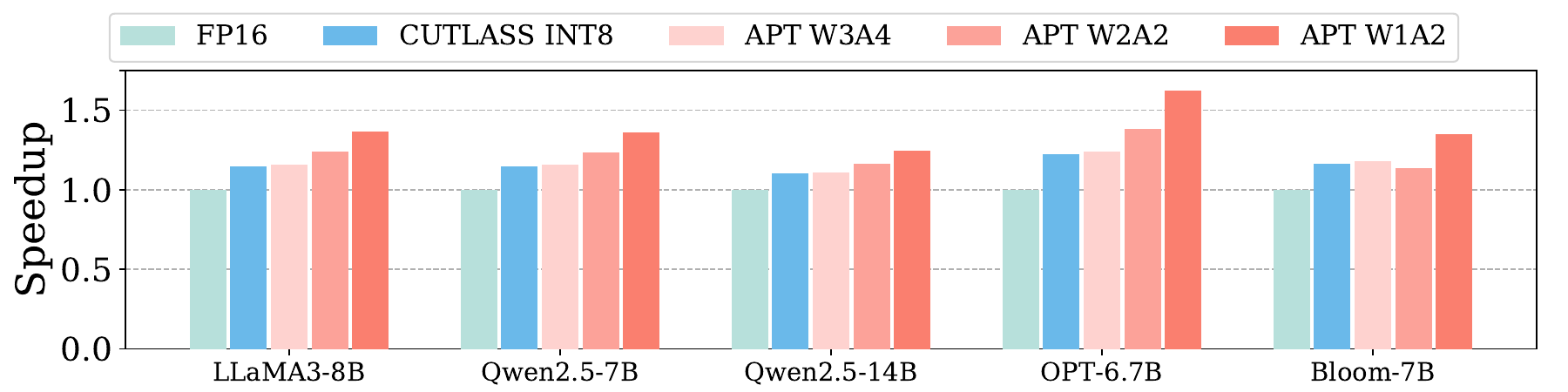}
        \label{fig:model-prefill-h800}
    }
    \subfigure[Decode on H800]{
        \includegraphics[width=0.48\linewidth]{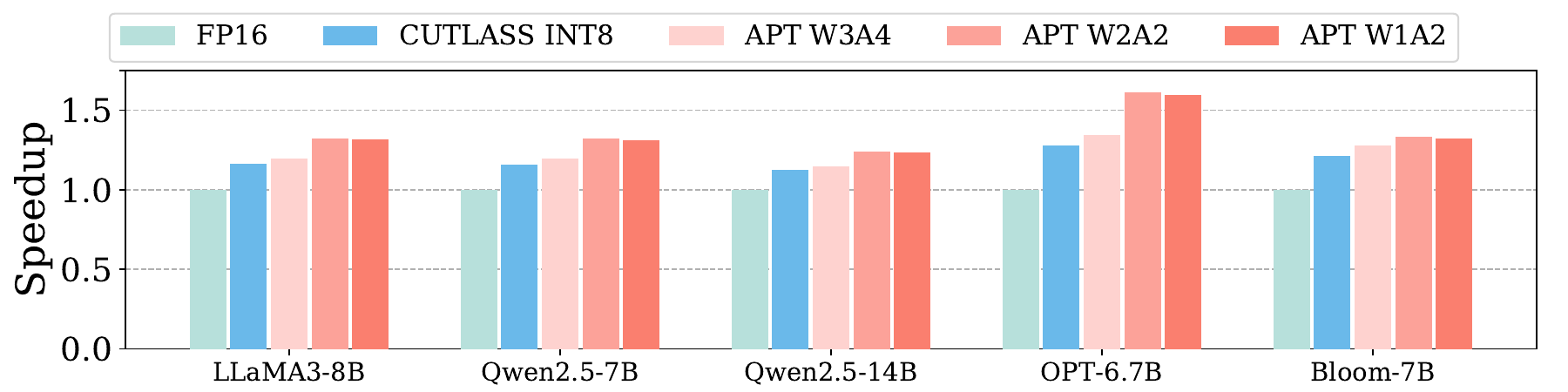}
        \label{fig:model-decode-h800}
    }
    \vspace{+1.5em}
    \caption{\rThreeMSB{Comparison of APT and other methods in accelerating LLM inference on different GPU platforms.}}
    \label{fig:model-eval-gpus}
    \vspace{-1.5em}
\end{figure*}

% \rThreeMSB{Considering the space limitations of the paper, 
\rThreeMSB{We \rTwoFC{further} conduct experiments only on a subset of models with larger parameter sizes. These models are more widely used and better demonstrate the acceleration benefits of APT-LLM. On RTX 4090 and NVIDIA H800, our method achieves significant speedups, reaching up to 2.44$\times$ over FP16 and 1.65$\times$ over CUTLASS integer baselines. Notably, the relative speedups on these platforms are slightly lower than those observed on RTX 3090. This trend can be attributed to two primary factors: the exponential improvement in baseline performance and a corresponding shift in performance bottlenecks.}

% \rThreeMSB{The raw computational power of new architectures, such as Ada Lovelace and Hopper, has dramatically increased, with next-generation \rTwoFC{TC}s accelerating FP16 computations. As baseline inference times become exceptionally fast, the potential for achieving high relative speedup diminishes. Additionally, the evolution of GPU hardware has shifted the performance bottleneck. On earlier GPUs like the RTX 3090, memory access and computation were the main contributors to inference time, and most acceleration methods, including APT-LLM, focused on optimizing these aspects. However, as memory bandwidth and computation have improved significantly, their relative impact on inference time has reduced, leading to lower relative speedup ratios compared to the RTX 3090. This outcome is a natural result of hardware progress and does not indicate that APT-LLM loses effectiveness. Instead, it adds an extra layer of acceleration, continuing to deliver significant speedups despite evolving GPU architectures.}
\rThreeMSB{The primary factor behind the reduced speedup is the significant hardware acceleration provided by newer architectures like Ada Lovelace (RTX 4090) and Hopper (H800), which drastically improve throughput. However, the data recovery phase in our method, which does not benefit from \rTwoFC{TC}s, has not scaled proportionally, resulting in diminished overall speedup. Moreover, with substantial advancements in memory bandwidth and computational capabilities, their influence on inference time has diminished. Hence, the proportion of time dedicated to other operations, such as quantization and dequantization, has increased, leading to reduced relative speedup ratios when compared to RTX 3090. This is a consequence of hardware improvements, and 
% while the relative speedup has decreased, 
APT-LLM continues to provide valuable acceleration even as GPU architectures evolve.}

\subsection{Kernel Mapping Evaluation}

In this section, we evaluate the acceleration impact of APT's kernel mapping on both MatMul kernels and overall model inference. First, we design an ablation study to demonstrate the performance contributions of memory scheduling and kernel mapping, the two key design optimizations in APT. Then, using the LLAMA3-8B model as a case study, we enumerate the kernel configurations adopted by APT under different precision settings across different inference phases.
% (prefill and decode).

\subsubsection{\textbf{Ablation Study}}

To further validate the effectiveness of our design choices, we conduct an ablation study to isolate the impact of key optimization strategies in APT. Specifically, we evaluate the contribution of our memory scheduling strategy and kernel mapping optimization by selectively disabling these components and measuring the resulting performance degradation. This experiment is conducted under the W1A2 configuration with a 64/4k/4k MatMul setup for comparison.

\begin{figure}[t]
	\centering
	\includegraphics[width=\linewidth]{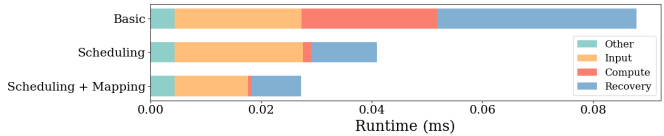}
	\caption{Ablation study demonstrating the impact of memory scheduling and kernel mapping on APT performance. `Basic' represents a naive implementation without optimizations, while `Scheduling' incorporates only memory scheduling without kernel mapping.}
	\label{fig:ablation}
        \vspace{-1.5em}
\end{figure}

Fig. \ref{fig:ablation} illustrates the impact of our optimization strategies. First, compared to the baseline implementation, memory scheduling does not reduce the time spent on input processing. However, it provides two key benefits: (1) the allocation of two separate shared memory blocks significantly hides computation latency, and (2) APT’s memory management strategy is designed for efficient data recovery, leading to a notable reduction in data recovery time. As a result, memory scheduling alone achieves a 2.15$\times$ speedup over the baseline.

Furthermore, incorporating kernel mapping optimizations accelerates all stages of computation. This is because different kernel configurations dynamically adjust the number of threads allocated for data loading, computation, and recovery, leading to a comprehensive performance improvement. Compared to the memory scheduling-only method, kernel mapping provides an additional 1.50$\times$ speedup, resulting in an overall 3.22$\times$ acceleration over the basic implementation.

\subsubsection{\textbf{Kernel Mapping in LLMs}}
\begin{figure}[tb]
	\centering
	\includegraphics[width=0.9\linewidth]{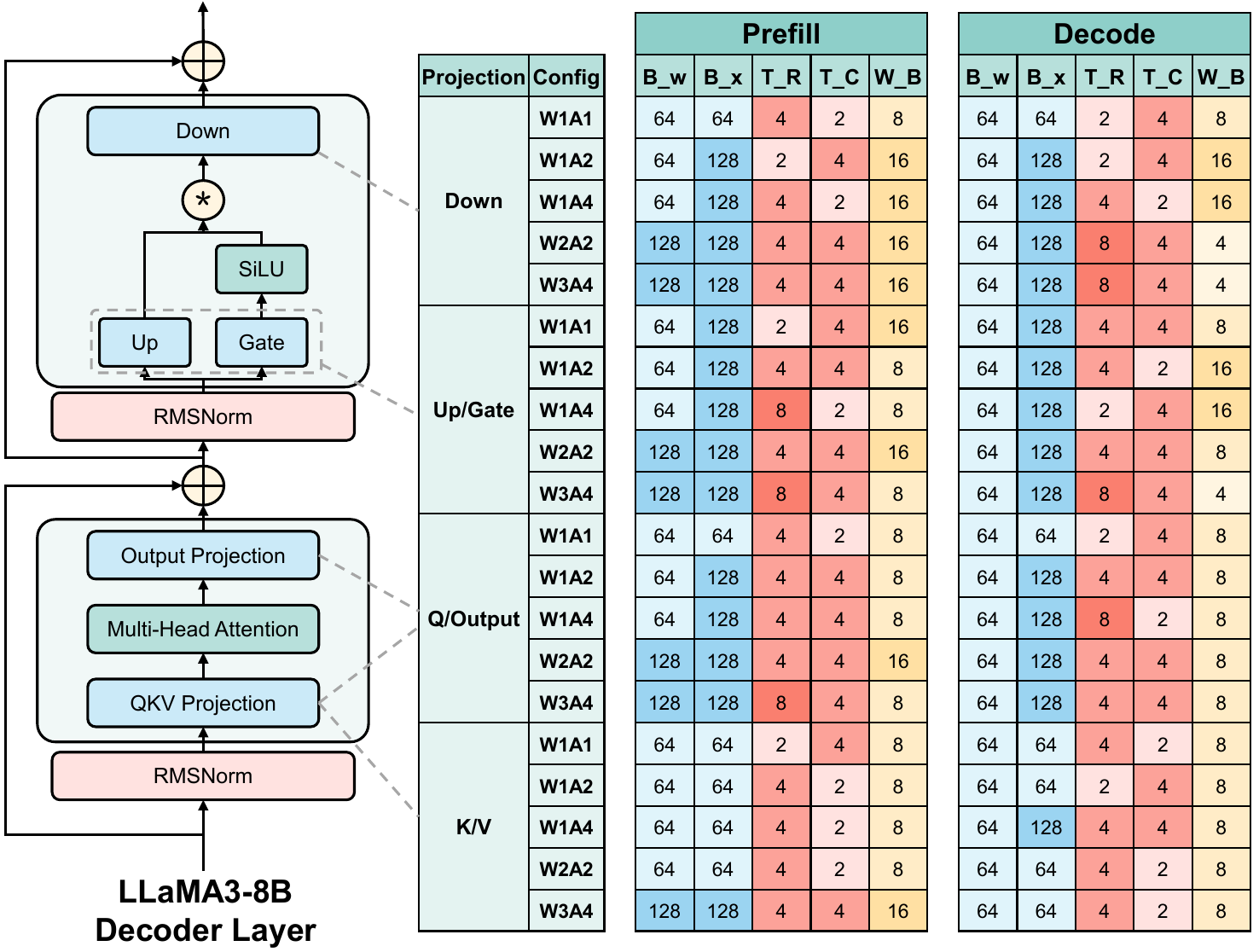}
        \vspace{1.5em}
	\caption{Kernel mapping hyperparameter configurations for different layers in an LLM under various activation and weight precision settings.}
	\label{fig:kernel mapping}
        \vspace{-1em}
\end{figure}

Fig. \ref{fig:kernel mapping} illustrates the optimal kernel configurations adopted by APT across different activation-weight precision settings during the prefill and decode phases of LLAMA3-8B inference. 
% Compared to Fig. \ref{fig:matmul different}, it is evident that APT dynamically adjusts its kernel configurations not only across different layers under the same precision setting but also based on varying precision levels and inference stages. This adaptability enables APT to optimize performance across diverse computational scenarios.
In the comparison of different configurations, we observe that varying bit-width configurations, different projection layers, and distinct computation stages (prefill or decode) all significantly impact the optimal configuration for APT. 
% For example, when the K/V projection uses W3A4 for inference, four out of the five variable hyperparameters in APT differ across the two phases.

A comparative analysis of different configurations reveals the complexity of formulating a universal heuristic for the manual adjustment of kernel configurations. This complexity is heightened in the context of differing LLM architectures and diverse model sizes, where kernel configurations require dynamic adaptation. Thus, manually optimizing these configurations proves to be almost impractical.

In contrast, the kernel mapping approach proposed in this work provides an effective solution by automatically identifying the optimal configuration for each layer within a single inference run. This method can even be extended to large-scale model inference scenarios where different layers operate at varying precision levels. 

\subsection{\rThreeMSB{Model Accuracy Evalution}}

\rThreeMSB{To verify that the APT-LLM method does not negatively impact the accuracy of the original quantized model, we apply APT-LLM to several open-source quantization methods\cite{smoothquant,omni,ashkboos2024quarot,atom}, quantizing them to W4A4. We then evaluate the perplexity of the Llama-2-7B model on the WikiText-2 dataset\cite{wikitext2}, comparing the performance of the model with and without APT-LLM during inference.}

\begin{table}[tb]
	\centering
	\caption{\rThreeMSB{WikiText-2 perplexity results on quantized LLAMA-2 models with and without APT-LLM.}}
        \vspace{+2.0em}
	\begin{tabular}{c|c|cc|cc}
        \hline
        \multirow{2}{*}{Method} & \multirow{2}{*}{Bits} & \multicolumn{2}{c|}{w/o APT-LLM} & \multicolumn{2}{c}{w/ APT-LLM} \\ \cline{3-6} 
                                &                       & 7B              & 13B            & 7B             & 13B           \\ \hline
        Baseline                & FP16                  & 5.47            & 4.88           & -              & -             \\ \hline
        SmoothQuant\cite{smoothquant}             & \multirow{2}{*}{W6A6} & 6.20            & 5.18           & 6.20           & 5.19          \\
        OmniQuant\cite{omni}               &                       & 5.87            & 5.14           & 5.88           & 5.13          \\ \hline
        SmoothQuant\cite{smoothquant}             & \multirow{3}{*}{W4A4} & 83.12           & 35.88          & 83.11          & 35.87         \\
        OmniQuant\cite{omni}               &                       & 14.26           & 12.30          & 14.26          & 12.31         \\
        QuaRot\cite{ashkboos2024quarot}                  &                       & 6.10            & 5.40           & 6.10           & 5.39          \\ \hline
        Atom-128G\cite{atom}               & \multirow{2}{*}{W4A4} & 6.03            & 5.26           & 6.03           & 5.26          \\
        QuaRot-128G\cite{ashkboos2024quarot}             &                       & 5.93            & 5.26           & 5.92           & 5.26          \\ \hline
        \end{tabular}
	\label{tab:ppl}
        \vspace{-1.5em}
\end{table}

\rThreeMSB{Table \ref{tab:ppl} 
% displays the results, showing 
shows that APT-LLM achieves perplexity scores nearly equivalent to those of the traditional quantization techniques. This finding suggests that \rTwoFC{our bipolar-INT format effectively maintains quantization performance.}}
% our method effectively maintains quantization performance while avoiding substantial numerical decline.}
\rThreeMSB{The slight discrepancies in perplexity observed are attributed to the intrinsic constraints of numerical representation in computing systems. Although our algorithm is mathematically lossless, it necessitates alterations in the quantization parameters of the original models, which are usually stored in FP32 or FP16 formats. Given that numerical representations lack absolute precision, these essential adjustments to the parameters unavoidably introduce errors to the perplexity of the original quantized model. Nonetheless, as our experiments indicate, these errors are confined within tolerable limits.}
% \rThreeMSB{This empirical validation with our theoretical proof of lossless conversion in \ref{subsec:bipolar_int}, shows that APT-LLM achieves computational acceleration without compromising model accuracy.}

\section{Conclusions} \label{sec:conclusions}
%In this paper, we propose APT-LLM, an efficient GPU acceleration scheme for arbitrary-precision large language models (LLMs) that tackles challenges of limited GPU Tensor Core (TC) support, inefficient memory management, and rigid kernel optimizations. 
\rTwoMSB{In this paper, we propose APT-LLM, an efficient GPU acceleration scheme for arbitrary-precision LLMs. APT-LLM tackles the challenges of limited GPU Tensor Core (TC) support, inefficient memory management, and rigid kernel optimizations.}
We first introduce the bipolar-INT data format that enhances parallel processing and allows for efficient and lossless conversion with signed INT. We then develop a bit-level MatMul technique that supports arbitrary precision and efficiently utilizes GPU TCs. Our scheme also incorporates a multi-level memory management system for accelerating data recovery, and a dynamic kernel mapping strategy for optimal performance across diverse matrix configurations in LLMs.
% \rOneFC{Comprehensive evaluations across various popular LLMs demonstrate APT-LLM achieves up to 3.99$\times$ speedup over FP16 baselines and 2.16$\times$  speedup over NVIDIA CUTLASS INT4 acceleration, showing promising potential for efficient LLM deployment in resource-constrained environments.}
\rThreeMSB{Comprehensive evaluations across various popular LLMs demonstrate APT-LLM achieves up to a 3.99$\times$ speedup compared to FP16 baselines and a 2.16$\times$ speedup over NVIDIA CUTLASS INT4 acceleration on the RTX 3090. On the RTX 4090 and H800, APT-LLM   achieves up to 2.44$\times$ speedup over FP16 and 1.65$\times$ speedup over CUTLASS integer baselines, showing promising potential for efficient LLM deployment in resource-constrained environments.}
% Experimental results demonstrate APT-LLM's effectiveness, achieving up to 3.99× speedup over FP16 baselines and 1.98× speedup over CUTLASS INT4 acceleration, showing promising potential for efficient LLM deployment in resource-constrained environments.}

\section*{Acknowledgment}

\rTwoFC{The authors would like to sincerely thank their reviewers for the constructive feedback, 
% Dr. Zhibin Wang for the helpful discussion, and Yuhao Ji for expanding the scope of ideas and providing valuable insights that greatly contributed to this work.
and Dr. Zhibin Wang and Yuhao Ji for the helpful discussion. % with valuable insights.
}

\normalem
\bibliographystyle{IEEEtran}
\bibliography{ref}

\end{document}